\documentclass[a4paper,10pt,journal,twoside]{IEEEtran}
\IEEEoverridecommandlockouts
\usepackage{fontspec}
\usepackage{amsmath,amssymb,amsfonts}
\usepackage{graphicx}
\usepackage{booktabs,tabularx,array,multirow,threeparttable}
\usepackage{stfloats}
\usepackage{url}
\usepackage[colorlinks=true,linkcolor=black,citecolor=black,filecolor=black,urlcolor=blue]{hyperref}
\usepackage{xcolor}
\usepackage{cite}
\renewcommand{\thesection}{\arabic{section}}
\renewcommand{\thesubsection}{\thesection.\arabic{subsection}}
\graphicspath{{}}
\hypersetup{pdftitle={InsertAnything: Generalizable Contact-Rich Precision Insertion from Simulation to Reality},pdfauthor={Zhenghua Ma et al.}}

\title{InsertAnything: Generalizable Contact-Rich Precision Insertion from Simulation to Reality}
\author{Zhenghua Ma$^{a,b,*}$, Xinpan Meng$^{a,b}$, Zeyu Liu$^{a,b}$, Muyuan Ma$^{a,b}$, Hengdi Zhang$^{c}$, Houcheng Li$^{a,b,*}$, and Long Cheng$^{a,b,*}$%
\thanks{Zhenghua Ma, Xinpan Meng, Zeyu Liu, Muyuan Ma, Houcheng Li, and Long Cheng are with the School of Artificial Intelligence, University of Chinese Academy of Sciences, Beijing 100049, China, and the State Key Laboratory of Multimodal Artificial Intelligence Systems, Institute of Automation, Chinese Academy of Sciences, Beijing 100190, China (e-mail: long.cheng@ia.ac.cn; lihoucheng2017@ia.ac.cn). Hengdi Zhang is with PaXini AI Technology (Beijing) Co., Ltd., Beijing 100080, China. *Corresponding authors.

We acknowledge PaXini AI (Beijing) Co., Ltd., whose tactile sensors were used in the real-robot experiments.}}

\begin{document}
\bstctlcite{IEEEcontrol}
\maketitle

\begin{abstract}
Contact-rich precision insertion is a key manipulation skill in robotic assembly. Tight clearances make insertion more sensitive to alignment errors and prone to collisions and jamming, while variations in geometry and clearance across parts further complicate policy reuse. We present a reinforcement learning framework that trains insertion policies entirely in simulation for direct deployment without real-world demonstrations or policy fine-tuning. By combining target poses with compact three-dimensional fingertip force feedback, the policy learns to search for alignment and correct its motion despite errors in the estimated hole position. A decoupled gated reward coordinates alignment and insertion. Force-signal smoothing and state-independent standard deviations stabilize the learning process. The resulting policies perform real-world insertion across multiple hole geometries with a minimum nominal clearance of 0.02 mm and improve success while reducing peak contact forces under hole-position errors. Cross-clearance and cross-geometry evaluations further confirm policy generalization. The system achieved the first perfect score of 20/20 on ManipulationNet's peg-in-hole benchmark under its Human-in-the-Loop protocol, with fully autonomous insertion motions. A single policy trained only on a simulated hexagonal insertion task achieved an overall success rate of 95.0\% across eight unseen real-world insertion tasks. These results show that learning entirely in simulation can yield precision insertion skills that can be deployed directly and reused across real-world tasks. The project website (\url{https://mzhsoul.github.io/InsertAnything/}) provides open-source simulation and real-robot experiment scripts, assets, and trained checkpoints.
\end{abstract}
\begin{IEEEkeywords}
Contact-rich insertion; Reinforcement learning; Sim-to-real transfer; Force feedback; Policy generalization
\end{IEEEkeywords}

\hypertarget{introduction}{%
\section{Introduction}\label{introduction}}

Accurately inserting a peg into a target hole is a fundamental capability in contact-rich robotic manipulation and a key step in industrial assembly, electrical connector insertion, and equipment maintenance \cite{luGeneralPurposeIntelligentAgent2020}. Intelligent manufacturing calls for robots that can flexibly manipulate different products and components \cite{zhou2018toward,day2018robotics,zhongIntelligentManufacturing2017}. Precision insertion, however, requires more than reaching a target position: the robot must continually adjust its position, orientation, and applied force during contact. When the mating clearance narrows to tens of micrometers, even a small error in the target pose can cause collisions with the hole rim and jamming, while continued motion along the insertion direction can increase contact force \cite{whitneyQuasiStaticAssembly1982,lozano-perezAutomaticSynthesisFinemotion1984}. Precision insertion therefore depends on correcting motion in response to actual contact, rather than simply improving positioning accuracy in free space.

Compliant control provides a foundation for such precision insertion tasks. Impedance control regulates the relationship between motion and force, whereas hybrid position/force control assigns motion and force objectives to different directions \cite{hoganImpedanceControl1985,raibertHybridPositionForce1981}. Combined with search motions and force feedback, these methods can improve insertion reliability under initial misalignment \cite{vanwykComparativePeginholeTesting2018}. Contact pose identification \cite{jinContactPoseIdentification2021}, active contact sensing \cite{kimActiveExtrinsicContact2022}, and compliant motion planning based on contact constraints \cite{chenRobustPeginholeAssembly2025} further use contact information to resolve uncertainty. Human-robot collaborative assembly also uses impedance control to combine human decisions with robot execution \cite{zhaoHumanRobotCollaboration2023}. Beyond regulating motion during contact, these methods require task-specific choices of search direction, alignment actions, and when to advance. Learning these decisions from data offers a way to reduce the effort of designing motion strategies for each task.

In reinforcement learning, a robot learns how to choose actions by interacting with its environment. Previous studies have achieved search and insertion at small clearances on real robots \cite{inoueDeepReinforcementLearning2017} and combined policy learning with variable impedance control \cite{luoReinforcementLearningVariable2019}. Parameterized insertion actions further organize exploration into action primitives with continuous parameters \cite{zhangLearningInsertionPrimitives2022}. To improve learning efficiency, demonstration-assisted reinforcement learning uses expert experience for insertion into sockets at varying positions and for a range of industrial assembly tasks \cite{vecerikPracticalApproachInsertion2019,luoRobustMultimodalPolicies2021}. Human-in-the-loop reinforcement learning combines demonstrations with online interventions to achieve high success rates in dexterous manipulation \cite{luoPreciseDexterousRobotic2025}. Diffusion Policy and action chunking have also improved the ability to learn complex motions from demonstrations \cite{chiDiffusionPolicyVisuomotor2023,zhaoLearningFineGrainedBimanual2023}. Real-world interaction and expert experience provide effective learning signals for these methods. Conducting this trial-and-error learning in simulation could further reduce data collection and real-world training costs for new tasks.

Efficient contact simulation has made this training approach possible. Factory and Orbit support parallel contact interaction and robot learning \cite{narangFactoryFastContact2022,mittalOrbitUnifiedSimulation2023}, while research on reward learning shows that how task progress is described affects the efficiency of learning contact skills \cite{wuLearningDenseRewards2021}. IndustReal combines geometric rewards, curriculum learning, and deployment control to transfer assembly policies directly from simulation to reality \cite{tangIndustRealTransferringContactrich2023}. FORGE further uses force feedback and force constraints to learn insertion, gear meshing, and nut threading under pose and dynamics uncertainty \cite{noseworthyFORGEForceguidedExploration2025}. To address differences in contact dynamics, online admittance residual learning adjusts compliance parameters through real-world interaction \cite{zhangEfficientSimtorealTransfer2023}. Dynamic compliance tuning instead combines desired-force planning with gain adjustment during execution to apply simulation-trained policies directly to tight clearances and different real-world insertion tasks \cite{zhangBridgingSimtorealGap2024}. These advances show that contact skills can transfer directly from simulation to real robots. They raise two related questions: how to achieve stable learning of motion corrections after contact when localization is biased, and whether the learned corrections remain effective when mating clearances and part geometries change.

The choice of feedback affects both what a policy can sense and how it transfers from simulation to reality. Multimodal representation learning has shown that vision and touch provide complementary information \cite{leeMakingSenseVision2019}. MimicTouch learns contact-rich manipulation from human tactile demonstrations \cite{yuMimicTouchLeveragingMultimodal2025}, while 3D-ViTac and ReTac-ACT support fine manipulation and precision assembly through three-dimensional visuotactile representations and state-gated fusion, respectively \cite{huang3DViTacLearningFinegrained2025,ruanReTacACTStategatedVisiontactile2026}. In addition to high-dimensional tactile representations, three-dimensional tactile sensors that distinguish normal and tangential loads provide robots with direct force measurements \cite{yangSoftTactileUnit2025,qiFlexible3DTactileElectronics2026}. In simulation, TACTO, differentiable tactile simulation, and TacSL have advanced the generation of tactile images and contact force fields, as well as policy learning \cite{wangTACTOFastFlexible2022,xuEfficientTactileSimulation2023,akinolaTacSLLibraryVisuotactile2025}. VT-Refine further combines real-world demonstrations with reinforcement learning in simulation to improve visuotactile assembly \cite{huangVTrefineLearningBimanual2025}. These approaches require models of sensor responses to contact; vision-based tactile sensing also involves deformation and imaging. For insertion tasks with calibrated target hole poses, a compact three-dimensional fingertip force signal offers another option: the policy can correct motion directly from changes in force without high-dimensional tactile observations. Learning must then accommodate signal fluctuations and differences between simulated and real sensor responses.

If a policy learns to adjust motion from contact feedback, the next question is whether this ability extends to parts not encountered during training. Tactile-RL and insertion methods based on visual representations of the peg-hole gap have demonstrated transfer to unknown geometries \cite{dongTactileRLInsertionGeneralization2021,xie2022LearningFill}. AutoMate uses expert policy distillation and simulation training to produce a generalist policy for multiple assemblies used in training \cite{tangAutoMate2024}. Meta-reinforcement learning and offline meta-reinforcement learning instead use prior experience to accelerate adaptation to new tasks \cite{schoettlerMetareinforcementLearningRobotic2020,zhaoOfflineMetareinforcementLearning2022}. These studies explore direct transfer, multitask learning, and adaptation to new tasks. We ask whether the same policy trained entirely in simulation on standard hole geometries can remain effective when the mating clearance, hole geometry, or real-world object changes, without retraining. Controlled simulation experiments can separate the effects of changes in clearance and geometry, while standardized physical assembly benchmarks provide consistent conditions for validating real-world transfer \cite{kimbleBenchmarkingProtocolsEvaluating2020,chenManipulationNetInfrastructureBenchmarking2026}. Table 1 compares the proposed framework with representative approaches.

\begin{table*}[t]
\centering
\caption{Training approaches, information used during execution, and real-world validation of representative insertion methods.}
\label{tab:1}
\scriptsize
\setlength{\tabcolsep}{2pt}
\begin{tabular}{@{}p{0.15\linewidth}p{0.22\linewidth}p{0.22\linewidth}p{0.35\linewidth}@{}}
\toprule
Method & Training approach & Execution inputs & Representative real-world validation \\
\midrule
Inoue et al. \cite{inoueDeepReinforcementLearning2017} & Real-world reinforcement learning & Position, force/torque & Circular-hole insertion at 20 μm clearance \\
Xie et al. \cite{xie2022LearningFill} & Policy learning in simulation; perception trained on real images & Peg-hole gap extracted from images & The same policy inserts into unseen hole geometries \\
IndustReal \cite{tangIndustRealTransferringContactrich2023} & Reinforcement learning in simulation & Pose & Direct real-world transfer of policies for multiple assembly tasks \\
AutoMate \cite{tangAutoMate2024} & Expert policies trained in simulation and combined & Pose, part geometry & One policy performs 20 assemblies used in training \\
Zhang et al. \cite{zhangBridgingSimtorealGap2024} & Supervised learning from simulation data & Pose, force/torque & 0.02 mm clearance; transfer to real connectors \\
FORGE \cite{noseworthyFORGEForceguidedExploration2025} & Reinforcement learning in simulation & Pose, three-dimensional end-effector force & Insertion, gear meshing, and nut threading; changes in part size \\
ReTac-ACT \cite{ruanReTacACTStategatedVisiontactile2026} & Learning from real-world demonstrations & Images, tactile observations & Insertion at three clearances, down to 0.1 mm \\
\textbf{Ours} & \textbf{Reinforcement learning entirely in simulation; no real-world fine-tuning} & \textbf{Pose, three-dimensional fingertip force} & \textbf{Multiple geometries at 0.02 mm clearance; transfer across clearances and geometries; one policy for eight unseen tasks} \\
\bottomrule
\end{tabular}
\end{table*}

To address these questions, the proposed framework combines target poses with fingertip force feedback to learn precision insertion policies in simulation for direct real-world deployment (Fig. 1). The target pose provides a nominal insertion goal, while fingertip force reflects actual contact, allowing the policy to correct motion as contact forces change. Building on this insight, a decoupled gated reward describes planar alignment, yaw alignment, and axial insertion separately and modulates the insertion reward according to the alignment state. Force-signal filtering and state-independent standard deviations stabilize learning from force feedback, while observation and dynamics randomization help the policy accommodate variations in sensing and execution conditions. The resulting policies can be deployed without real-world demonstrations or fine-tuning. We first evaluate insertion performance under hole-position errors, then examine whether these contact-based corrections remain effective when clearances, hole geometries, and real-world parts change.

\begin{figure*}[t]
\centering
\includegraphics[width=0.82\textwidth]{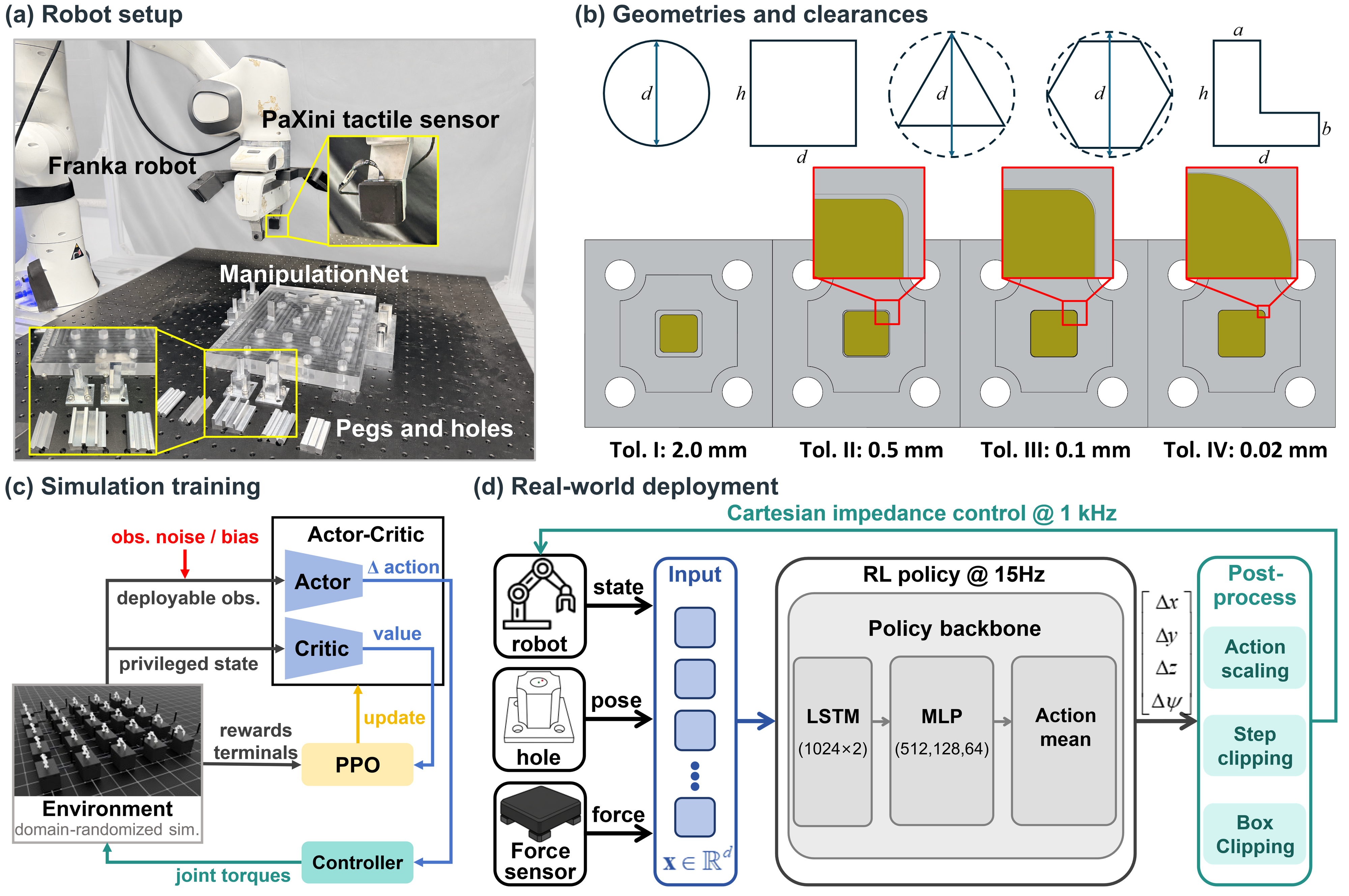}
\caption{Overview of learning precision insertion in simulation and deploying it on a real robot. (a) The robot platform with fingertip force sensors, shown alongside standard insertion workpieces and the ManipulationNet Peg-in-Hole Assembly benchmark board \cite{chenManipulationNetInfrastructureBenchmarking2026}. (b) Five hole geometries and four clearance levels of the standard insertion workpieces fabricated for this study. The bottom row illustrates the square-hole task, with close-up views showing progressively tighter clearances. (c) Asymmetric actor–critic training in simulation: the actor uses deployable observations, the critic additionally uses simulation state to estimate value, and PPO updates the policy under observation and dynamics randomization. (d) Real-world deployment: the trained policy generates mean actions from the robot state, calibrated target, and fingertip force. Postprocessing converts these actions into end-effector pose commands executed by a Cartesian impedance controller, without real-world demonstrations or policy fine-tuning.}
\label{fig:fig01}
\end{figure*}

The main contributions are as follows:

\begin{enumerate}
\def\labelenumi{\arabic{enumi}.}
\item
  A simulation-only framework combines nominal target poses with compact three-dimensional fingertip force feedback for precision insertion. It enables contact-aware motion correction and direct deployment to real robots without real-world demonstrations or policy fine-tuning.
\item
  A decoupled gated reward coordinates planar alignment, yaw alignment, and axial insertion, while force-signal smoothing and state-independent standard deviations stabilize policy learning. The resulting force-feedback policies improve robustness to hole-position errors and reduce peak contact forces.
\item
  Evaluations demonstrate generalization across clearances and geometries, with a single simulation-trained policy achieving 95.0\% success across eight unseen real-world insertion tasks. The system also achieves the first perfect 20/20 score on ManipulationNet's peg-in-hole benchmark under its Human-in-the-Loop protocol \cite{manipulationNetPegLeaderboard}.
\end{enumerate}

The remainder of this paper is organized as follows. Section 2 presents the learning and deployment methods. Section 3 reports learning performance, real-world precision insertion, and policy generalization, in that order. Section 4 discusses the main findings, and Section 5 concludes the paper.

\hypertarget{methods}{%
\section{Methods}\label{methods}}

\hypertarget{problem-formulation-and-robot-control}{%
\subsection{Problem formulation and robot control}\label{problem-formulation-and-robot-control}}

We study precision insertion of a grasped object into a target hole, as shown in Fig. 1(a). The object is already held by the gripper at the start of each insertion. The robot progressively aligns the object by correcting its horizontal position and rotation about the insertion axis, and continues to adjust its motion using contact feedback during insertion. An insertion is considered successful when the object reaches the prescribed depth and satisfies the alignment criteria for the task. The target hole pose is directly available in simulation and obtained through calibration on the real robot. This pose provides a nominal reference for insertion, while fingertip force feedback allows the policy to adjust motion based on actual contact.

The training environments are built in Isaac Lab using Factory \cite{narangFactoryFastContact2022,nvidiaIsaacLabGPUaccelerated2025}, with five standard shapes: circular, square, hexagonal, triangular, and L-shaped. Each shape has four clearance levels, denoted Tol. I--IV from loosest to tightest. The standard insertion workpieces designed and fabricated for this study have corresponding nominal mating clearances of 2, 0.5, 0.1, and 0.02 mm, respectively (Fig. 1(b)). These standard tasks are used for policy training and provide a basis for the subsequent evaluations of cross-clearance and cross-geometry generalization.

The insertion task is formulated as a Markov decision process (MDP). Let \(\mathbf s_t\), \(\mathbf a_t\), and \(r_t\) denote the state, policy action, and immediate reward at decision step \(t\), respectively. The policy observation \(\mathbf o_t\) consists of information available on the real robot: the robot's fingertip position and orientation relative to the target hole, the estimated end-effector linear and angular velocities, the three-dimensional fingertip force, and the smoothed action from the previous step. On the real robot, \(\mathbf o_t\) is constructed using hand-guided calibration, robot proprioception, and fingertip force sensors. Policy learning uses the asymmetric actor--critic architecture shown in Fig. 1(c) \cite{pintoAsymmetricActorCritic2018}. The actor generates actions from \(\mathbf o_t\). During training, the critic estimates value from the simulation state \(\mathbf s_t\), which includes robot and workpiece states, friction coefficients, and other information. The resulting value estimates provide a more informative learning signal for policy updates. We denote the policy by \(\pi_\theta(\mathbf a_t\mid\mathbf o_t)\) and the value function by \(V_\phi(\mathbf s_t)\), with parameters \(\theta\) and \(\phi\), respectively. The policy is trained to maximize the expected discounted return:

\[
\mathcal J(\theta)
=\mathbb E_{\pi_\theta}\left[\sum_{t=0}^{T-1}\gamma^t r_t\right],
\tag{1}
\]
where \(T\) is the episode length and \(\gamma\) is the discount factor. The policy is updated using proximal policy optimization (PPO) \cite{schulmanProximalPolicyOptimization2017}. PPO uses advantage estimates to increase the probability of beneficial actions and clipping to curb overly large policy updates.

The policy outputs a six-dimensional normalized action \(\mathbf a_t\) specifying three-dimensional translational and rotational increments of the end effector. Clipping, temporal smoothing, and scaling convert these increments into a target end-effector pose. During execution, roll and pitch are fixed to keep the gripper pointed downward, leaving four effective degrees of freedom: three translations and yaw adjustment. A low-level compliant controller tracks this target pose, as shown in the real-world deployment workflow in Fig. 1(d). Implementation details are provided in Supplementary Material S1.

In simulation, the compliant controller computes a PD control action in operational space from the Cartesian pose error of the end effector and maps it to joint torques using the Jacobian transpose \cite{khatibOperationalSpace1987}:

\[
\boldsymbol\tau_t
=\mathbf J_t^{\mathsf T}\mathbf w_t+\boldsymbol\tau_{\mathrm{null},t},
\qquad
\mathbf w_t
=\mathcal D_\delta\left(\mathbf K_p\mathbf e_t-\mathbf K_d\mathbf v_t\right).
\tag{2}
\]

In Eq. (2), \(\boldsymbol\tau_t\) is the joint torque command, \(\mathbf w_t\) is a six-dimensional Cartesian control vector comprising forces and torques, and \(\mathbf J_t\) is the robot's geometric Jacobian. The vector \(\mathbf e_t\) contains position and orientation errors, and \(\mathbf v_t\) contains the end-effector linear and angular velocities. The matrices \(\mathbf K_p\) and \(\mathbf K_d\) are the stiffness and damping matrices, respectively. The operator \(\mathcal D_\delta\) applies a componentwise dead zone with threshold \(\delta\), and \(\boldsymbol\tau_{\mathrm{null},t}\) is the null-space torque used to regulate the robot's posture through its redundant degrees of freedom. On the real robot, the Cartesian impedance control interface in SERL tracks the target poses supplied by the policy \cite{luoSERLSoftwareSuite2024}. Compliant tracking allows contact to influence the actual motion, while the policy continuously adjusts the target pose.

\hypertarget{reward-design-for-decoupling-alignment-and-insertion}{%
\subsection{Reward design for decoupling alignment and insertion}\label{reward-design-for-decoupling-alignment-and-insertion}}

With very small clearances, even slight planar or yaw errors can cause the peg to press against the hole rim, so moving downward alone does not necessarily produce effective insertion. Existing assembly methods use distances between keypoints arranged along the insertion axis to describe three-dimensional positional errors between the workpiece and its target, without accounting for yaw errors in noncircular shapes \cite{narangFactoryFastContact2022,noseworthyFORGEForceguidedExploration2025}. Planar alignment, yaw alignment, and vertical insertion are decoupled to better reflect the geometric constraints of precision insertion, as shown in Fig. 2.

Let \(d_{xy}\) denote the planar distance between the peg reference point and its final target, \(d_z\) the vertical distance between them, and \(d_R\) the yaw error. Each error is mapped to a bounded shaping reward:

\[
r_i(d_i)=\frac{1}{\exp(k_i d_i)+b_i+\exp(-k_i d_i)},
\qquad i\in\{xy,R,z\}.
\tag{3}
\]

The terms \(r_{xy}\), \(r_R\), and \(r_z\) represent the planar, yaw, and vertical reward components, respectively. The parameter \(k_i>0\) controls how rapidly the reward decays as the error increases, and \(b_i\geq0\) adjusts its peak value. We apply a planar gate \(m_{xy}\) and a yaw gate \(m_R\) to the vertical reward:

\[
m_{xy}(d_{xy})=\exp(-\kappa d_{xy}),
\qquad
m_R(d_R)=\mathbb I[d_R<\epsilon_R],
\tag{4}
\]
where \(\kappa>0\) is the decay coefficient of the planar gate, \(\epsilon_R\) is the yaw-gating threshold, and \(\mathbb I[\cdot]\) is the indicator function. As the planar error decreases, the gate weight increases continuously, so the same reduction in vertical error earns a larger insertion reward. The yaw gate enables this reward once the angular error falls within the specified range. The planar and yaw rewards remain active throughout, guiding the policy to continue correcting alignment during insertion. For circular pegs, we omit the yaw reward and set \(m_R=1\).

The complete reward at each step combines these geometric terms with action penalties and insertion-stage bonuses:

\[
\begin{aligned}
r_t={}&w_{xy}r_{xy}(d_{xy})+w_Rr_R(d_R)\\
&+w_zm_{xy}(d_{xy})m_R(d_R)r_z(d_z)\\
&-\lambda_a\|{\tilde{\mathbf a}}_t\|_2
-\lambda_{\Delta a}\|{\tilde{\mathbf a}}_t-{\tilde{\mathbf a}}_{t-1}\|_2\\
&+w_e I_{\mathrm{eng},t}+w_h I_{\mathrm{half},t}\\
&+w_s I_{\mathrm{succ},t}.
\end{aligned}
\tag{5}
\]

The coefficients \(w_{xy}\), \(w_R\), and \(w_z\) weight the geometric reward components. The action \({\tilde{\mathbf a}}_t\) has undergone the clipping and temporal smoothing described in Section 2.1. The weights \(\lambda_a\) and \(\lambda_{\Delta a}\) penalize action magnitude and changes in action, respectively, discouraging excessively large or abrupt actions. The indicators \(I_{\mathrm{eng},t}\), \(I_{\mathrm{half},t}\), and \(I_{\mathrm{succ},t}\) each equal one when the corresponding insertion-depth condition is met at the current step and zero otherwise, with weights \(w_e\), \(w_h\), and \(w_s\), respectively. Detailed reward parameters are provided in Supplementary Material S1.

Keypoint rewards couple alignment and insertion, encouraging vertical motion before adequate alignment and making very tight-clearance tasks difficult to learn. The proposed reward addresses this problem by coordinating alignment and insertion through gating. Section 3.1 evaluates the effectiveness of this design through reward comparisons.

\begin{figure}[t]
\centering
\includegraphics[width=0.96\columnwidth]{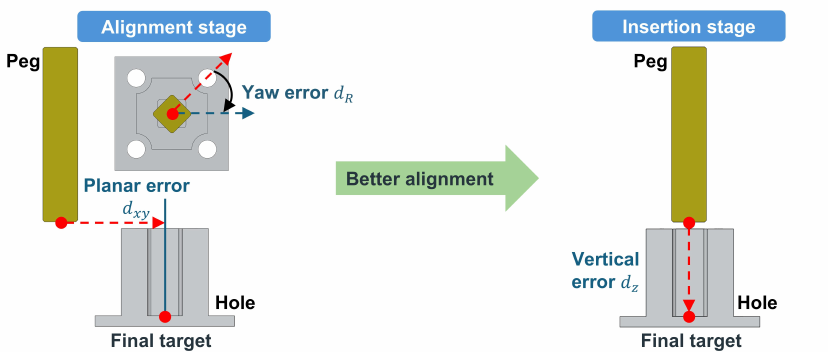}
\caption{Decoupled reward design. Planar error $d_{xy}$, yaw error $d_R$, and vertical error $d_z$. As alignment improves, gating increases the reward for approaching the final insertion target.}
\label{fig:fig02}
\end{figure}

\hypertarget{learning-with-fingertip-force-feedback}{%
\subsection{Learning with fingertip force feedback}\label{learning-with-fingertip-force-feedback}}

The reward design provides learning objectives for alignment and insertion, while execution must also account for errors in the target hole pose. When the calibrated position differs from the actual hole position, continued motion toward the nominal target may cause the peg to press against the hole rim. Fingertip forces change with actual contact and robot motion, allowing the policy to adjust its motion in response. With this feedback, the policy can continue searching for an effective insertion direction despite position errors. It can also reduce excessive contact forces from sustained downward pushing to prevent peg tilting or slipping. Stable learning with force feedback is supported by combining force-signal smoothing with a state-independent exploration scale, as shown in Fig. 3.

Training uses the simulated gripper--peg contact force. On the real robot, three-dimensional readings from the fingertip force sensor are expressed in the force-coordinate convention used during training. Before each insertion, we record a force baseline while the peg is not in contact with the target hole and subtract it from subsequent readings to remove the initial force bias. The force signal can also fluctuate rapidly during initial contact and sliding exploration. An exponential moving average (EMA) smooths the baseline-subtracted force:

\[
\bar{\mathbf f}_t
=\alpha\mathbf f_t+(1-\alpha)\bar{\mathbf f}_{t-1},
\tag{6}
\]
where \(\mathbf f_t\) denotes the baseline-subtracted three-dimensional force, \(\bar{\mathbf f}_t\) is the smoothed force, and \(\alpha\in(0,1]\) is the smoothing coefficient, set to 0.25. The smoothed force is provided to the policy as part of the observation \(\mathbf o_t\). The left side of Fig. 3 illustrates how EMA smooths transient spikes and rapid fluctuations, allowing the policy to use changes in contact while reducing the effects of short-lived disturbances on policy learning.

Force signals affect not only actions but also the exploration scale in Gaussian policies with state-dependent standard deviations. In these insertion tasks, this parameterization is associated with fluctuations in the exploration scale and a decline in policy performance late in training. We therefore retain an action mean that depends on force feedback and use learnable, state-independent standard deviations:

\[
\pi_\theta(\mathbf a_t\mid\mathbf o_t)
=\mathcal N\!\left(
\boldsymbol\mu_\theta(\mathbf o_t),
\operatorname{diag}(\boldsymbol\sigma_\theta^2)
\right).
\tag{7}
\]

The policy outputs the action mean \(\boldsymbol\mu_\theta(\mathbf o_t)\). The vector \(\boldsymbol\sigma_\theta\) contains the standard deviations for each action dimension, represented by learnable log-standard-deviation parameters. Each action dimension has its own parameter; the same set of parameters is shared across states and updated during PPO training. The right side of Fig. 3 compares the two parameterizations: force feedback still affects action selection through the mean, but changes in the current observation do not directly alter the exploration scale.

This distinction can also be expressed through the conditional differential entropy of the Gaussian policy:

\[
\mathcal H\!\left[\pi_\theta(\cdot\mid\mathbf o_t)\right]
=\frac{d}{2}\log(2\pi e)
+\sum_{j=1}^{d}\log\sigma_{\theta,j}.
\tag{8}
\]

The entropy \(\mathcal H\) measures the randomness of the policy's action distribution before postprocessing, \(d=6\) is the policy output dimension, and \(\sigma_{\theta,j}\) is the standard deviation of the \(j\)th action dimension. For a given set of policy parameters, Eq. (8) does not vary with the current observation; during training, the standard deviations can still adjust to the exploration scale required by the task. EMA mitigates short-lived spikes and fluctuations in the force input, while state-independent standard deviations keep these fluctuations from directly determining the exploration scale. Section 3.1 analyzes their effects on learning through ablation experiments, with related mathematical analysis provided in Supplementary Material S2.

During real-world deployment, insertion is executed using the action mean output by the policy, with force feedback continuing to guide motion corrections. The training randomization described in the next section further accommodates variations in observations and contact during deployment.

\begin{figure}[t]
\centering
\includegraphics[width=0.96\columnwidth]{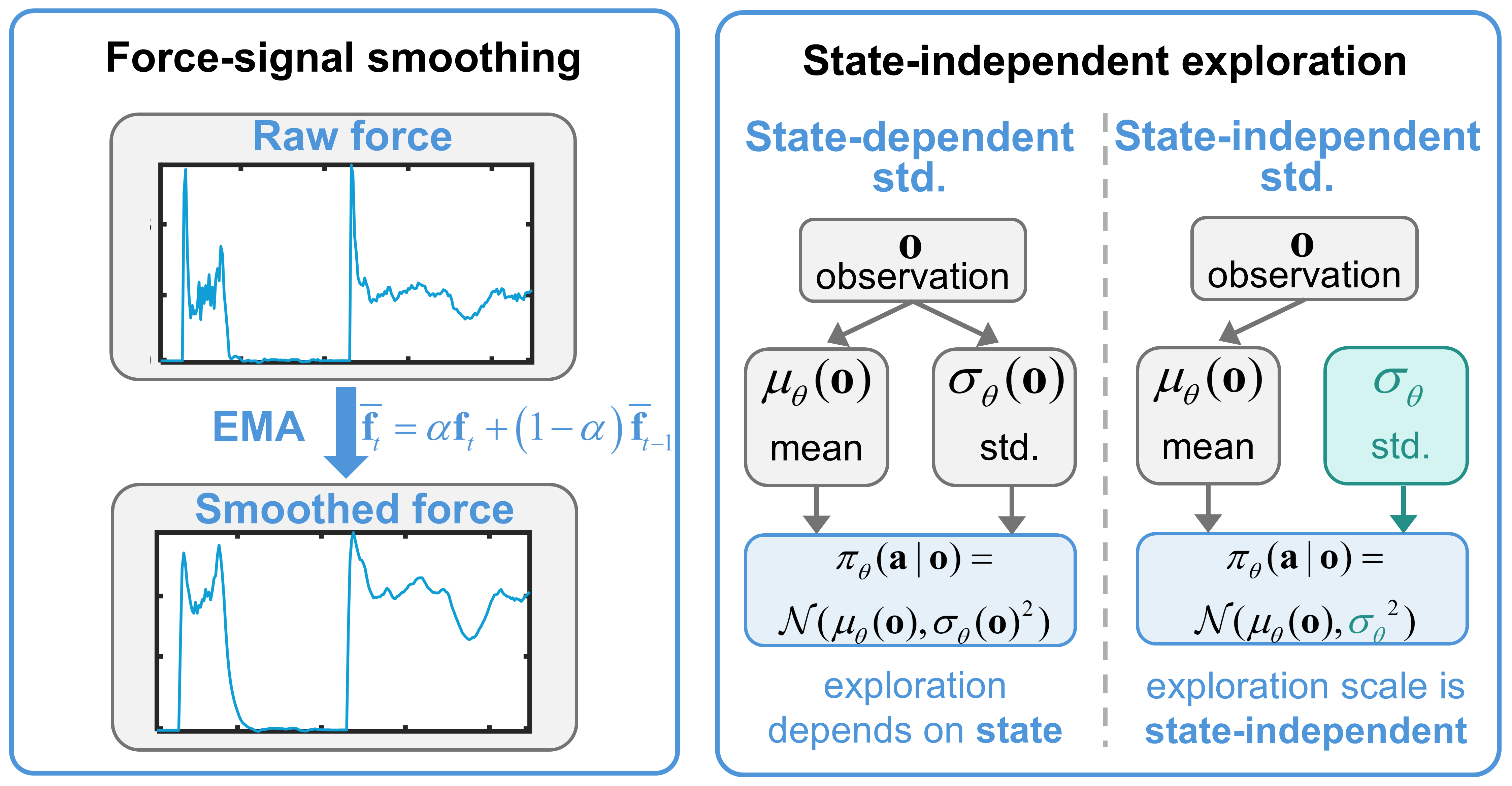}
\caption{Fingertip force-signal smoothing and state-independent exploration. The left side shows the baseline-subtracted force before and after EMA smoothing. The right side compares state-dependent and state-independent standard deviation parameterizations; std. denotes standard deviation. The policy predicts the action mean from the observation, while the standard deviations for each action dimension are shared across states and updated during training.}
\label{fig:fig03}
\end{figure}

\hypertarget{randomization-and-real-world-deployment}{%
\subsection{Randomization and real-world deployment}\label{randomization-and-real-world-deployment}}

Observation and dynamics randomization during training (Fig. 1(c)) helps insertion policies learned in simulation accommodate errors and variations during real-world execution. The initial poses of the target hole and gripper are randomized in each episode so that insertion begins with different position and orientation errors. Observation errors include persistent calibration biases and measurement noise that varies over time. Biases are added to the target hole position and end-effector orientation observations and remain fixed within each episode. Gaussian noise is added to position, orientation, estimated velocity, and force observations at each decision step. The policy thus learns motion corrections when the observed position does not exactly match the actual contact position, while accommodating sensor measurement noise.

Motion after contact also depends on friction and the robot's control response. Following the principle of dynamics randomization \cite{pengSimtorealTransferRobotic2018}, we randomize the friction coefficients of the peg and hole, the translational and rotational components of the compliant controller's stiffness matrix \(\mathbf K_p\), and the dead-zone threshold \(\delta\) at the start of each episode, keeping them fixed throughout the episode. Friction variations affect resistance after contact, while changes in stiffness and dead zones alter the control effort generated by pose commands. Training under these conditions allows the policy to learn how the same pose correction produces different motions under different contact and control conditions.

For the force input, random scaling and dropout supplement additive noise, exposing the policy to variations in force magnitude and availability during training. Specifically, a nonnegative scaling factor is sampled for each episode and applied uniformly to the three-dimensional force vector. With a given probability, the force input is also set to zero for the entire episode. Scaling preserves the force direction but changes its magnitude; when the force input is zeroed, the policy must still complete the task using pose and motion information. This design is intended to allow the policy to use force feedback for motion corrections while reducing its dependence on a particular force magnitude. The randomization parameters are provided in Supplementary Material S1.

The real-robot platform uses a Franka Emika arm with Paxini PX-6AXGEN3 tactile sensors mounted on the gripper fingertips. After training, the policy is deployed through the process shown in Fig. 1(d). The nominal target hole pose is obtained through hand-guided calibration. Robot proprioceptive states and fingertip forces are used to construct the policy input \(\mathbf o_t\) with the coordinate conventions, observation ordering, and normalization used during training; force signals are processed as described in Section 2.3. The policy updates its action mean at 15 Hz. The postprocessing described in Section 2.1 converts this mean into a target end-effector pose, which a Cartesian impedance controller tracks at 1 kHz. The random noise, force scaling, and dropout used during training are disabled at deployment, when the policy uses actual measured feedback. Policy learning takes place entirely in simulation, and real-world execution requires neither demonstration data nor policy fine-tuning. Deployment settings are provided in Supplementary Material S4.

\hypertarget{experiments-and-results}{%
\section{Experiments and Results}\label{experiments-and-results}}

The evaluation begins with ablation experiments and comparisons with other control methods in simulation. It then assesses real-world precision insertion and robustness to hole-position errors before examining how policies trained on source tasks generalize to changes in clearance, geometry, and the real-world objects being inserted.

\hypertarget{learning-effective-precision-insertion-policies}{%
\subsection{Learning effective precision insertion policies}\label{learning-effective-precision-insertion-policies}}

We first compare the decoupled gated reward with a keypoint reward and its two-stage curriculum on Hexagon-III to test whether the proposed reward can guide the policy to coordinate alignment and insertion. Each setting uses three training seeds and includes fingertip force observations. As shown in Fig. 4(a), training directly toward the final insertion target (the bottom of the hole) with the keypoint reward yields a success rate that remains near zero throughout training. The first curriculum stage reduces the insertion depth by raising the target position (Curr. I). The second stage continues training the resulting policy toward the final insertion target (Curr. II) and achieves a final-window success rate of 96.5\%. The decoupled gated reward uses the final target throughout and reaches a final-window success rate of 98.2\%. These results show that the decoupled reward can provide more effective guidance for learning without a curriculum. Its advantage is more pronounced when the policy does not use force observations, as shown in Supplementary Material S3.

The next experiment examines the roles of force-signal smoothing (EMA) and state-independent standard deviations (SI std.) on Square-IV, which has a tighter clearance. Figure 4(b) compares four settings: None, EMA, SI std., and EMA + SI std., all using fingertip force observations and the same decoupled reward. Policy performance declines late in training when neither design is used. EMA and state-independent standard deviations each improve learning, and combining them further improves the final-window success rate and overall learning efficiency (Table 2). These results support their complementary roles: EMA mitigates spikes and fluctuations in the force input, while state-independent standard deviations prevent contact forces from directly changing the exploration scale, supporting stable learning of motion corrections based on force feedback.

\begin{table}[t]
\centering
\caption{Ablation of force-signal EMA and state-independent standard deviations (SI std.) on Square-IV.}
\label{tab:2}
\scriptsize
\setlength{\tabcolsep}{2pt}
\begin{tabular}{@{}l>{\centering\arraybackslash}p{16mm}>{\centering\arraybackslash}p{22mm}>{\centering\arraybackslash}p{14mm}@{}}
\toprule
Setting & Final-window success rate ↑ & Training epoch to reach 50\% success ↓ & Normalized AUC ↑ \\
\midrule
None & 0.348 & — & 0.370 \\
EMA & 0.555 & 215.3 & 0.538 \\
SI std. & 0.777 & 325.7 & 0.533 \\
EMA + SI std. & \textbf{0.842} & \textbf{179.0} & \textbf{0.667} \\
\bottomrule
\end{tabular}
\end{table}

Note: The final-window success rate is the mean over the last 5\% of records; normalized AUC is the mean training success rate over equally spaced records. The epoch to reach 50\% success is the first epoch at which this threshold is reached. Only two seeds reach the threshold for None, so the mean epoch is not reported. Detailed definitions are provided in Supplementary Material S3.

After evaluating the method designs, we compare the learned policies with direct compliant insertion (Direct) \cite{whitneyQuasiStaticAssembly1982}, Archimedean spiral search (Spiral) \cite{vanwykComparativePeginholeTesting2018}, rectangular raster search (Raster) \cite{marvelMultiRobotAssembly2018}, and hybrid force/position control (Hybrid) \cite{liHybridForcePosition2025}. The first three methods and pose-only PPO use pose and motion information, while Hybrid and force-feedback PPO additionally use the same three-dimensional force feedback. Figure 4(c) compares these six methods on circular and L-shaped holes at Tol. I--IV under the same evaluation conditions. Both PPO variants are trained at Tol. III for the corresponding shape, and each policy is evaluated across all four clearance levels. The baseline control laws and evaluation settings are provided in Supplementary Material S3.

Across the eight tasks, the four traditional controllers achieve overall success rates of 20.0\%--29.4\%, compared with 71.5\% for pose-only PPO and 91.9\% for force-feedback PPO. Success rates generally decrease as clearances tighten, while the learned policies maintain relatively high success rates. On Circle-IV, force-feedback and pose-only PPO achieve 93.5\% and 53.6\%, respectively; on L-shape-IV, they achieve 68.2\% and 41.7\%. Under pose errors, reinforcement learning policies that continually adjust motion from feedback are therefore better suited to precision insertion than methods based on predefined search trajectories or fixed force-to-position mappings. Force feedback also provides useful information for motion correction at tight clearances. The next section further examines these capabilities on a real robot.

\begin{figure*}[t]
\centering
\includegraphics[width=0.84\textwidth]{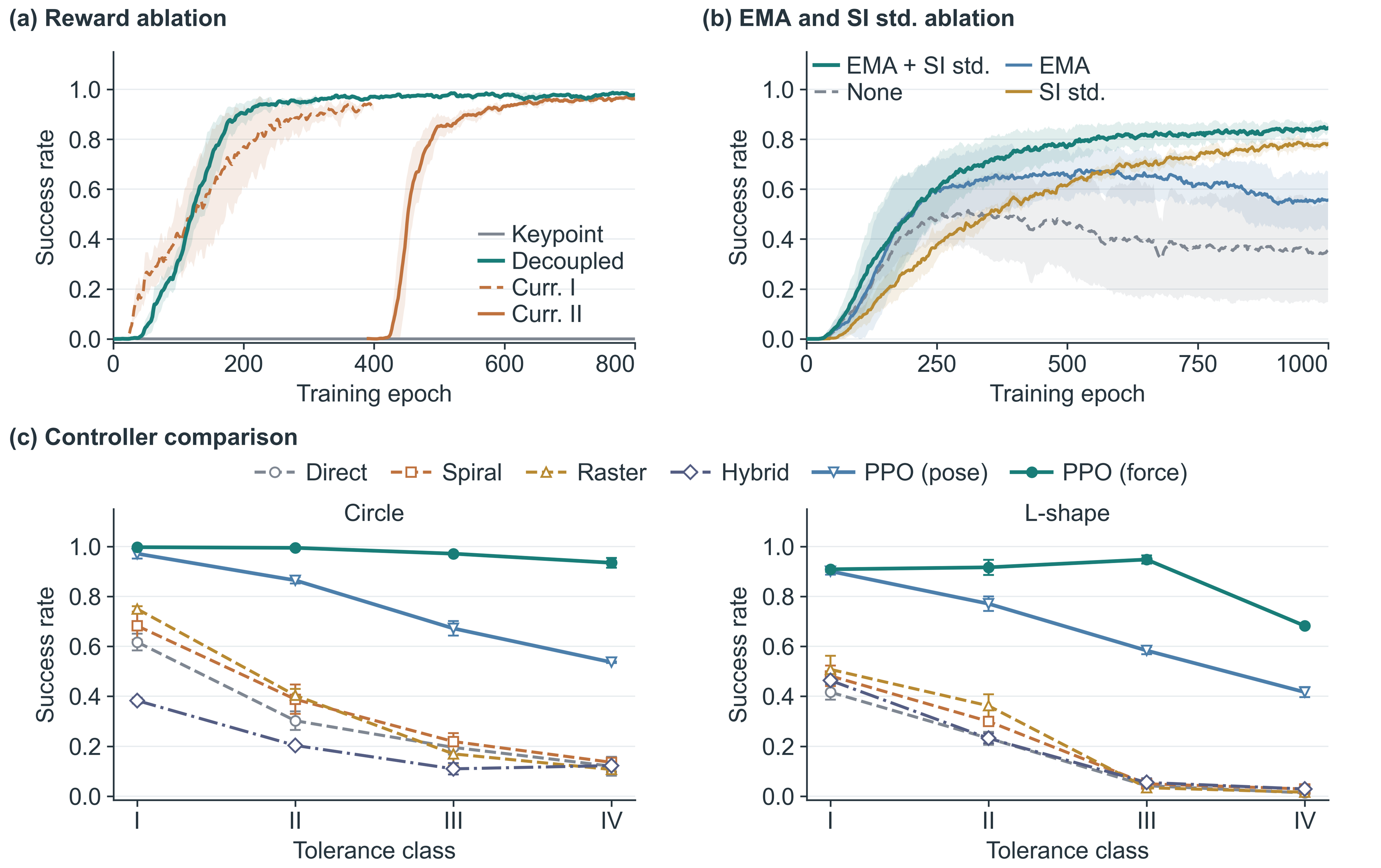}
\caption{Learning ablations and method comparisons. (a) Reward comparison on Hexagon-III with fingertip force feedback. Keypoint is a keypoint reward \cite{narangFactoryFastContact2022,noseworthyFORGEForceguidedExploration2025} augmented with a yaw-error reward term, and Decoupled is the proposed reward. For the keypoint reward, Curr. I trains with a raised target, and Curr. II continues training from the policy obtained in the first stage toward the final insertion target. (b) EMA and SI std. ablations on Square-IV. SI std. denotes learnable, state-independent standard deviations. None uses neither design, EMA and SI std. each use only the corresponding design, and EMA + SI std. uses both. Curves and shaded regions in (a) and (b) represent the mean and standard deviation of success rates across three training seeds, respectively. (c) Controller comparison under standard observation noise and dynamics randomization, with circular holes on the left and L-shaped holes on the right. Direct, Spiral, Raster, and Hybrid denote direct compliant insertion, Archimedean spiral search, rectangular raster search, and hybrid force/position control using three-dimensional force feedback, respectively. PPO (pose) and PPO (force) are independently trained policies without and with force observations, respectively. Points and error bars represent the mean and standard deviation across three evaluation seeds.}
\label{fig:fig04}
\end{figure*}

\hypertarget{real-world-precision-insertion-and-robustness-to-position-errors}{%
\subsection{Real-world precision insertion and robustness to position errors}\label{real-world-precision-insertion-and-robustness-to-position-errors}}

Real-world tight-clearance insertion is evaluated at Tol. IV on four types of workpieces fabricated for this study: circular, hexagonal, triangular, and L-shaped. For each geometry, the corresponding simulation-trained policies are deployed directly on the real robot. Each setting is tested in 20 trials, with the initial position and yaw of the grasped peg randomized and no additional bias injected into the observed hole position. The policies with force feedback achieve 20/20 successful insertions on both circular and hexagonal tasks, and 17/20 and 18/20 on triangular and L-shaped tasks, respectively. In comparison, the pose-only policies achieve 15/20, 10/20, 14/20, and 6/20 on the circular, hexagonal, triangular, and L-shaped tasks, respectively. Corrections learned with force feedback in simulation thus transfer to real-world insertion tasks and improve precision insertion success.

To further evaluate performance under hole-localization errors, we introduce a static bias in the observed planar position of the hole center in the real-robot Hexagon-IV task. The bias magnitude falls within three ranges: 0--1, 1--2, and 2--3 mm. Each policy is tested in 15 trials per range, with both policies using the same set of hole-position offsets. As shown in Fig. 5(a), force feedback increases the overall success rate from 68.9\% to 84.4\% while reducing the mean peak contact force from 3.08 N to 2.65 N. In the largest bias range of 2--3 mm, the policy with force feedback also achieves a higher success rate and lower peak forces (Table 3), indicating that the policy can correct localization errors while reducing excessive contact loads.

\begin{table}[t]
\centering
\caption{Real-robot insertion performance on Hexagon-IV with biased hole-position observations.}
\label{tab:3}
\scriptsize
\setlength{\tabcolsep}{2pt}
\begin{tabular}{@{}p{17mm}p{22mm}>{\centering\arraybackslash}p{16mm}>{\centering\arraybackslash}p{18mm}@{}}
\toprule
Bias range & Policy & Successful/total trials ↑ & Mean peak contact force (N) ↓ \\
\midrule
All (0–3 mm) & Pose-only PPO & 31/45 & 3.08 \\
All (0–3 mm) & PPO with force feedback & \textbf{38/45} & \textbf{2.65} \\
Largest bias (2–3 mm) & Pose-only PPO & 7/15 & 3.25 \\
Largest bias (2–3 mm) & PPO with force feedback & \textbf{11/15} & \textbf{2.71} \\
\bottomrule
\end{tabular}
\end{table}

To understand how force feedback helps correct hole-position errors, we further evaluate the simulated L-shape-III task with a +2 mm bias in each of the observed x and y coordinates of the hole center. The comparison uses the final planar positions of the peg after each policy runs for the same duration from the same grid of starting positions. As shown in Fig. 5(b), the final peg positions are more scattered with the pose-only policy, with some still tending toward the biased observed hole center. With force feedback, the final peg positions are more concentrated near the true hole center. The mean reduction in the peg's distance to the true hole center relative to its starting position increases from 1.98 mm to 4.96 mm. This result shows that fingertip force provides information about actual contact, allowing the policy to adjust insertion motions when the observed hole position is biased rather than relying solely on the observed target position.

\begin{figure*}[t]
\centering
\includegraphics[width=0.86\textwidth]{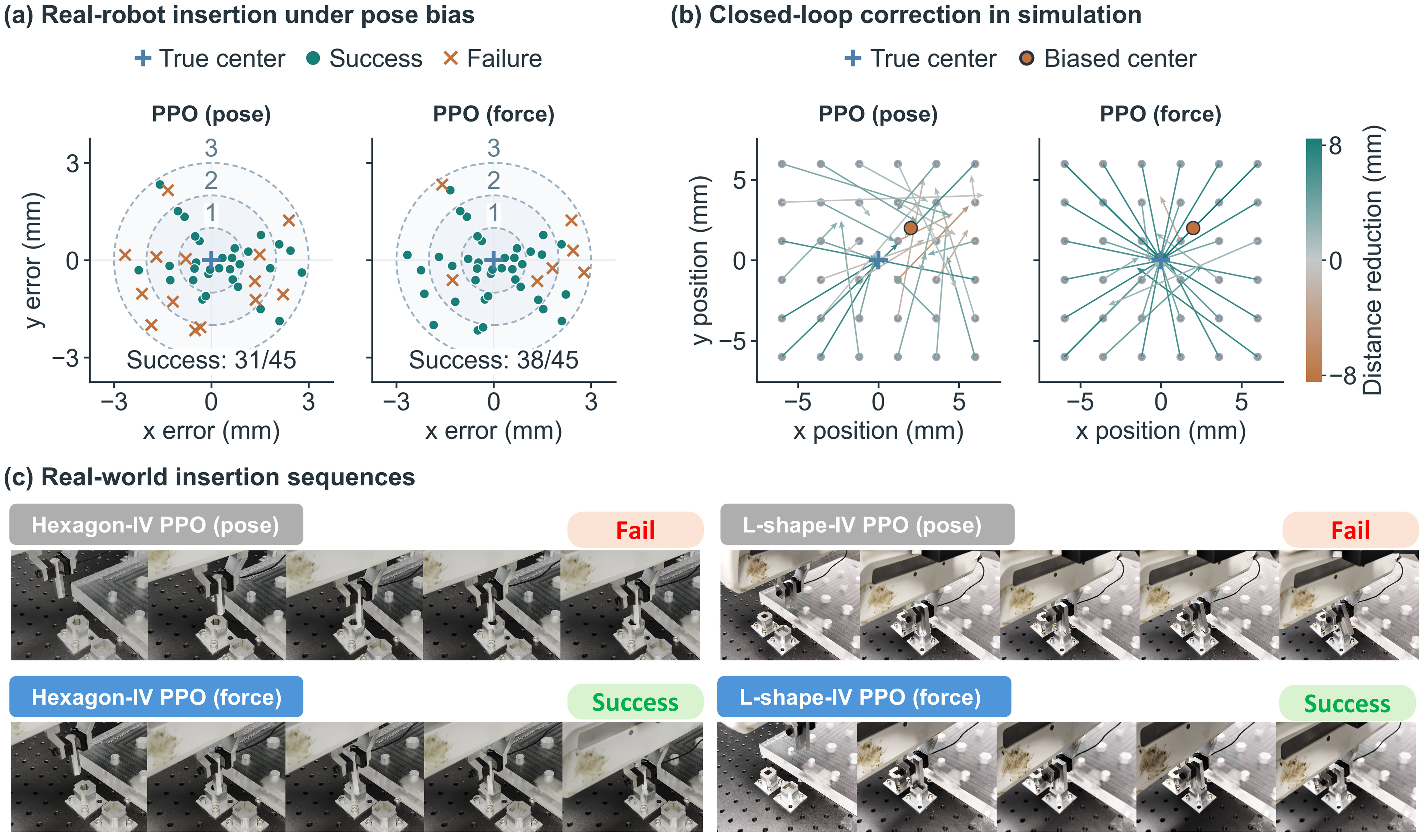}
\caption{Force feedback improves precision insertion under position errors. PPO (pose) and PPO (force) denote policies without and with force feedback, respectively. (a) Real-robot Hexagon-IV trials with biased hole-position observations. Each point's position indicates the bias in the observed hole center; green circles and orange crosses indicate success and failure, respectively. The dashed circles have radii of 1, 2, and 3 mm. Each policy is tested in 45 trials. (b) Closed-loop correction fields for L-shape-III in simulation, with a +2 mm bias in each of the observed x and y coordinates of the hole center and no other observation noise. Gray points mark 36 nominal peg starting positions arranged on a 6 × 6 grid, and arrows point to the final positions after 150 control steps. Color indicates the reduction in planar distance to the true hole center, with positive values indicating motion toward it. Blue crosses denote the true hole center in both (a) and (b); orange circles in (b) denote the biased observed hole center. (c) Real-world insertion sequences for Hexagon-IV and L-shape-IV, ordered chronologically from left to right within each row. The upper rows show alignment failure and peg tilting with the pose-only policies; the lower rows show successful insertion with force feedback.}
\label{fig:fig05}
\end{figure*}

The real-world insertion sequences in Fig. 5(c) illustrate two typical failure modes of the pose-only policies. In the hexagonal task, the policy fails to overcome the position error and achieve alignment. In the L-shaped task, excessive contact forces during exploration cause the peg to tilt severely, leading to insertion failure. The corresponding policies with force feedback successfully insert the pegs. Representative real-world insertions and the hexagonal comparison are shown in Supplementary Video 1, and the L-shaped insertion comparison is shown in Supplementary Video 2. These results show that policies trained entirely in simulation can use fingertip force feedback to adjust their motions during actual contact, improving insertion success under localization errors while reducing peak contact forces. The next section examines how the source policies perform when clearances and hole geometries change, and validates real-world cross-clearance transfer on the ManipulationNet benchmark.

\hypertarget{generalization-across-clearances-and-geometries}{%
\subsection{Generalization across clearances and geometries}\label{generalization-across-clearances-and-geometries}}

To examine whether insertion policies can be used under conditions beyond those encountered during training, we conduct controlled simulation experiments on five standard hole geometries. For each geometry, the selected policies use force feedback and are trained at Tol. III and Tol. IV. Each policy is evaluated at other clearances of the same geometry or on other geometries at the same clearance, with its training geometry and clearance serving as the in-distribution (ID) reference. Each test condition uses 128 parallel environments with three evaluation seeds, 0, 42, and 2026, for a total of 384 trials.

Figure 6(c) summarizes performance across transfer categories. The mean success rate for transfer across clearances within the same geometry is 92.4\%, close to the in-distribution rate of 93.8\%, showing that an insertion policy learned at one clearance can accommodate variations in clearance. The mean success rate for cross-geometry transfer is 70.6\%. Policies trained at Tol. III and Tol. IV and transferred to other geometries at the corresponding clearance achieve 77.9\% and 63.3\%, respectively. Compared with changing the clearance alone, changing the hole geometry requires the policy to handle different alignment and contact constraints, making transfer performance more dependent on the combination of source and target geometries.

Figures 6(a) and 6(b) show these differences in more detail. The Hexagon-III policy achieves a mean success rate of 86.3\% on the other four geometries, reaching 95.6\% and 95.1\% on circular and square holes, respectively. In the Tol. IV cross-geometry tests, the policy trained on the square geometry achieves the highest mean success rate. Policies trained on the triangular geometry transfer relatively poorly at both clearance levels, despite in-distribution success rates above 95\%. Transfer performance also depends on direction: at Tol. III, the policy trained on L-shaped holes achieves a 75.3\% success rate on triangular holes, whereas transfer in the reverse direction achieves 32.6\%. Learning high-precision insertion on the training task and reusing the resulting policy on other hole geometries are therefore distinct capabilities; the source geometry influences the range of other geometries to which the policy can adapt.

To examine the role of force feedback in cross-geometry transfer, we further compare the circular and L-shaped policies with force feedback, the same policies with their force observations set to zero, and independently trained pose-only policies. Both source and target tasks use Tol. III. As shown in Fig. 6(d), the policy trained on circular holes achieves a 75.3\% success rate on L-shaped holes with force feedback retained; this falls to 40.6\% when its force observations are zeroed. The independently trained pose-only policy achieves only 3.6\% on the same transfer task. By comparison, zeroing the force observations of the same circular policy on its in-distribution task reduces the success rate only from 97.1\% to 95.8\%. For transfer from L-shaped to circular holes, the policy with force feedback also outperforms the pose-only policy, with success rates of 90.6\% and 68.8\%, respectively. Force feedback also helps improve policy generalization to unseen hole geometries.

\begin{figure*}[t]
\centering
\includegraphics[width=0.84\textwidth]{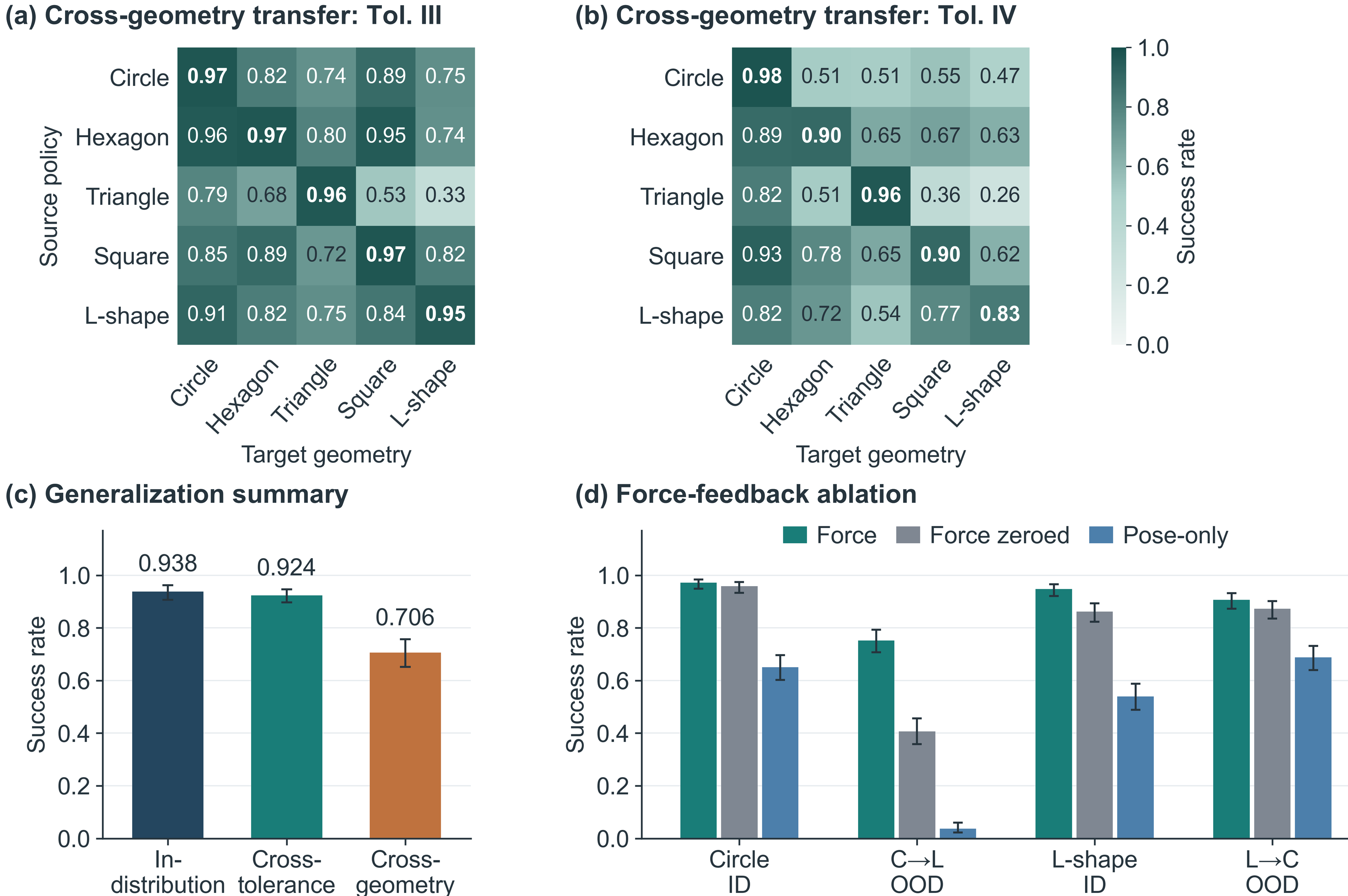}
\caption{Policy generalization across standard hole geometries and the role of force feedback. (a)(b) Cross-geometry success-rate matrices for Tol. III and Tol. IV. Rows indicate the geometry on which the source policy was trained, and columns indicate the target geometry; source and target tasks use the same clearance. Bold diagonal values indicate in-distribution performance. (c) Mean success rates for in-distribution tasks, cross-clearance transfer, and cross-geometry transfer, with equal weight assigned to each test condition. Cross-clearance transfer keeps the geometry unchanged, whereas cross-geometry transfer keeps the clearance unchanged. (d) Ablation of the effect of force feedback on policy generalization. Force denotes the standard force-feedback condition, Force zeroed sets the force observations of the same policy to zero, and Pose-only denotes an independently trained policy without force feedback. C and L denote circular and L-shaped holes, respectively, and arrows indicate the transfer direction; ID denotes in-distribution tasks, and OOD denotes unseen hole geometries. Error bars indicate 95\% confidence intervals.}
\label{fig:fig06}
\end{figure*}

The simulation experiments above evaluate policy transfer under changes in clearance and geometry. Generalization across clearances is further tested on ManipulationNet's real-world peg-in-hole assembly benchmark \cite{chenManipulationNetInfrastructureBenchmarking2026}. The benchmark includes five hole geometries, each with four clearance levels, Tol. I--IV, for a total of 20 insertion tasks (Fig. 7(a)). Their nominal mating clearances are 3, 1, 0.1, and 0.02 mm, respectively. The first two levels differ from the 2 and 0.5 mm clearances of the workpieces fabricated for this study, while the last two are the same. The pegs and task boards used for evaluation are manufactured under the supervision of the U.S. National Institute of Standards and Technology (NIST) and supplied to participating teams. Uniform dimensional and manufacturing specifications provide comparable physical test conditions across teams.

This evaluation uses five policies, one for each hole geometry, trained on the corresponding simulated Tol. III task. Each policy is applied directly to all four real-world clearance levels of its geometry, without real-world training or policy fine-tuning. Peg grasping and transfer are handled by motion planning, while insertion motions are performed entirely by the learned policy. Figure 7(b) shows real-world insertions for circular and L-shaped holes at all four clearance levels. The system successfully completed all 20 tasks in the official evaluation, earning a perfect score of 20/20. It was officially recognized as the first autonomous precision assembly system to achieve this score under the benchmark's Human-in-the-Loop protocol \cite{manipulationNetPegLeaderboard}. Under this protocol, insertion motions are performed fully autonomously by the robot, and the human operator only reports task status to the system. The official leaderboard entry and ManipulationNet's public announcement are reproduced in Fig. S3 of the Supplementary Material. This result extends cross-clearance capability from simulation to a standardized real-world benchmark: for each geometry, a policy trained at a single clearance can handle all four real-world clearance levels, from loose to extremely tight fits. Continuous execution is shown in Supplementary Video 1. The next section examines whether one of these source policies can be used for multiple unseen real-world insertion tasks with different structures.

\begin{figure*}[t]
\centering
\includegraphics[width=0.80\textwidth]{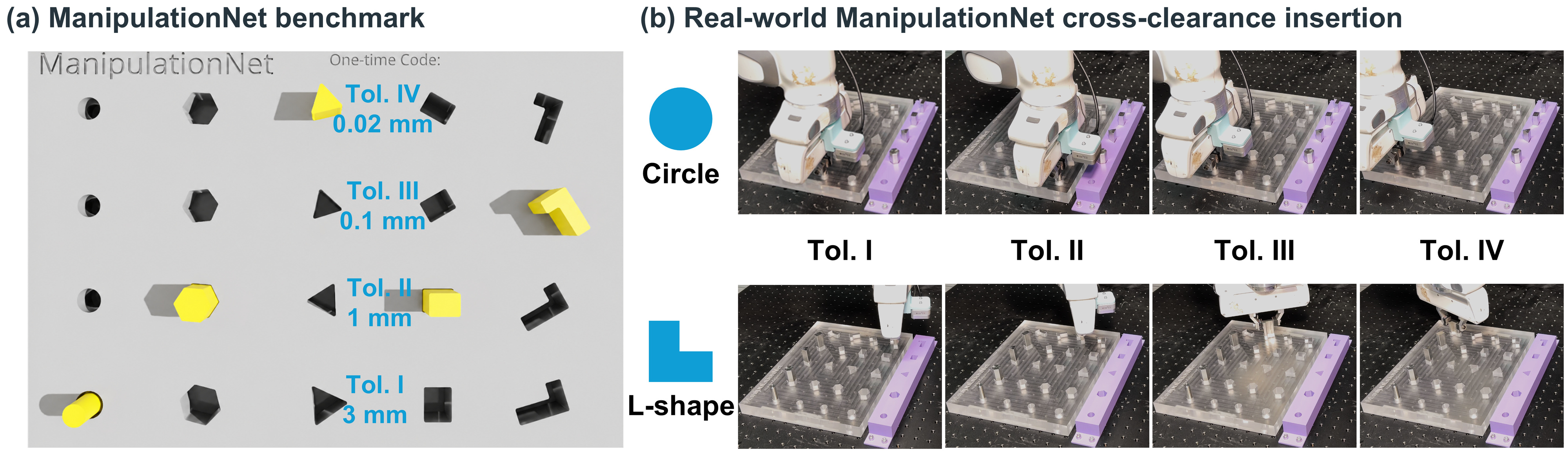}
\caption{Real-world cross-clearance insertion on the ManipulationNet benchmark. (a) Standardized peg-in-hole assembly board with circular, hexagonal, triangular, square, and L-shaped holes. Each geometry has four clearance levels, Tol. I-IV, with a minimum nominal clearance of 0.02 mm. (b) Examples of real-world insertion into circular and L-shaped holes.}
\label{fig:fig07}
\end{figure*}

\hypertarget{generalization-to-unseen-real-world-insertion-tasks}{%
\subsection{Generalization to unseen real-world insertion tasks}\label{generalization-to-unseen-real-world-insertion-tasks}}

Beyond cross-clearance and cross-geometry generalization among standard hole geometries, we further examine whether a source policy can be applied directly to real-world insertion tasks involving objects with different structures. In the simulated Tol. III cross-geometry tests in Section 3.3, the Hexagon-III policy with force feedback achieves the highest mean success rate on the other hole geometries. This policy is therefore selected for evaluation on eight unseen real-world insertion tasks. None of these objects was used to train the policy. All eight tasks use the same policy trained only on the simulated hexagonal Tol. III task, without fine-tuning. The experiments retain the same observation processing and settings for real-world deployment.

As shown in Fig. 8, the tasks include insertion of USB Type-A connectors, two-pin plugs, three-pin plugs, and DC power connectors, gear meshing, and insertion of three industrial connectors from electric vehicle battery circuits. The latter three have different interface geometries and are designated Industrial connector A, B, and C. These objects have mating structures that differ from those of standard peg-in-hole tasks. USB insertion requires alignment in a specific orientation, and two-pin and three-pin plugs require multiple blades to enter the corresponding holes simultaneously. DC connectors involve coaxial mating of an internal pin and an outer barrel. The gear task requires the teeth to mesh during insertion. The electric vehicle connectors further extend the evaluation to mating interfaces in actual industrial components.

Each task is evaluated over 20 trials, with 152 successful insertions out of 160 trials across the eight tasks, yielding an overall success rate of 95.0\%. Industrial connector A, B, and C achieve 20/20, 19/20, and 20/20 successful insertions, respectively; the counts for all tasks are shown in Fig. 8. Representative executions are presented in Supplementary Video 1. The same policy trained entirely in simulation on a standard peg-in-hole insertion task can therefore directly insert a range of real-world connectors and mesh gears, demonstrating its ability to generalize to unseen insertion tasks.

\begin{figure*}[t]
\centering
\includegraphics[width=0.84\textwidth]{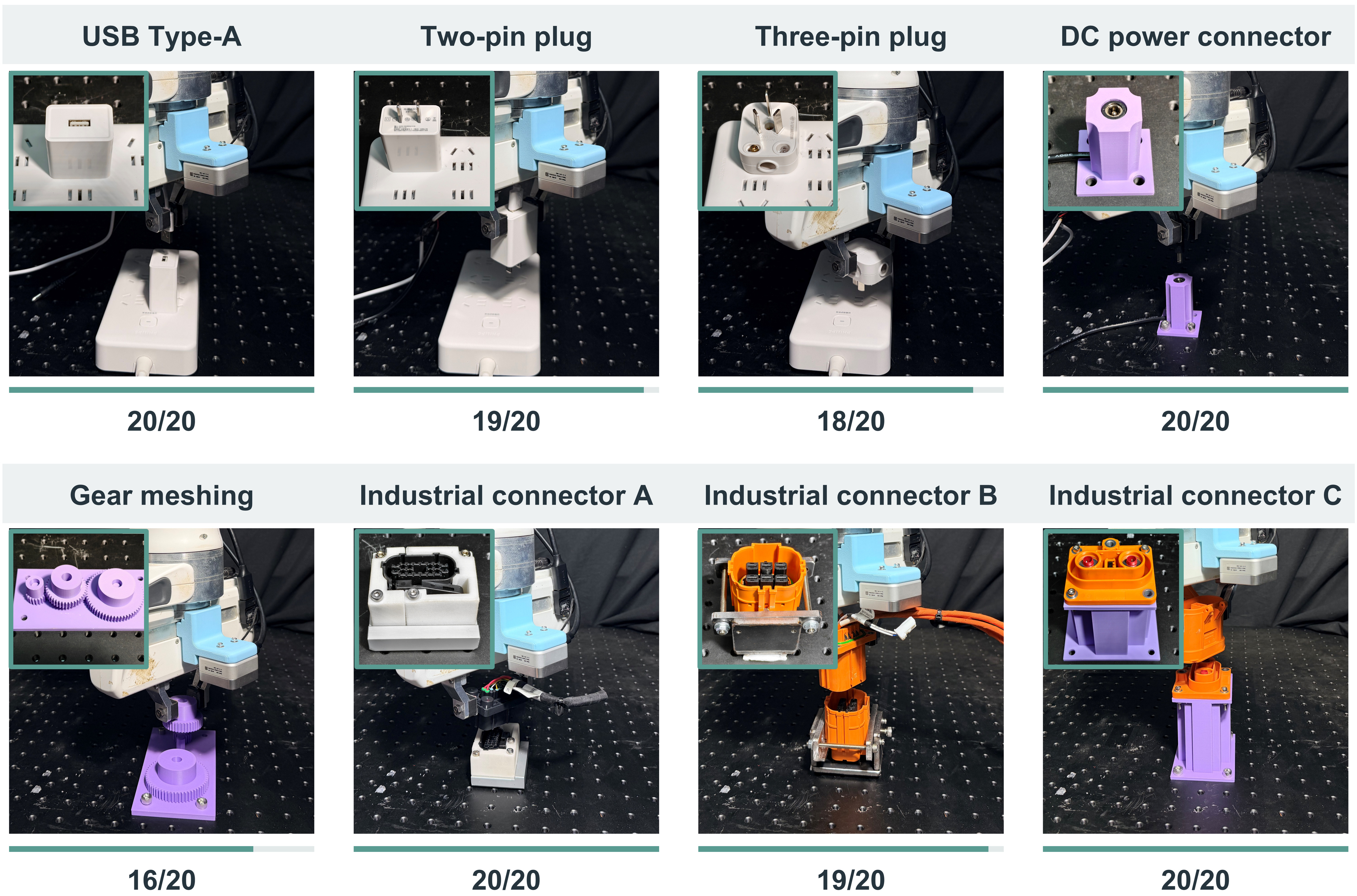}
\caption{Performance of the same Hexagon-III policy on eight unseen real-world insertion tasks. Each panel shows a representative real-world image for the corresponding task. The numbers below indicate successful insertions/total trials, and the thin teal bars indicate success rates. The top row shows, in order, a USB Type-A connector, a two-pin plug, a three-pin plug, and a DC power connector; the bottom row shows gear meshing and Industrial connector A, B, and C from electric vehicle battery circuits.}
\label{fig:fig08}
\end{figure*}

\hypertarget{discussion}{%
\section{Discussion}\label{discussion}}

Reliable precision insertion depends not only on target localization but also on how the robot adjusts its motion after contact. The target pose provides a nominal geometric reference, while fingertip forces reflect contact in real time, allowing the policy to continue correcting alignment when the observed hole position is biased. Figure 5 provides complementary evidence. The simulated displacement fields show that the force-feedback policy moves the peg closer to the true hole center. The real-robot experiments with biased hole-position observations show improved insertion success together with reduced peak contact forces. These findings show that force feedback helps the policy identify effective correction directions under localization errors, improving insertion success while reducing peak contact forces. This feedback-based correction was learned in simulation and applied directly to real-world tight-clearance tasks, demonstrating the value of compact three-dimensional force feedback in overcoming localization errors and achieving precision insertion.

The generalization experiments further show that insertion policies learned on standard hole geometries can be reused under mating conditions outside the training distribution. Performance across clearances shows that the policies can adapt to changes in clearance, while transfer performance from different source geometries provides a basis for selecting reusable policies. Experiments that zero the force observations of the same policy further support the role of force feedback in adapting to unseen hole geometries (Fig. 6): contact information obtained during execution helps the policy respond to changes in target geometry. This reuse extends from standard profiles to physical parts with different structures, as shown by real-world cross-clearance execution on ManipulationNet and the selected Hexagon-III policy's success in tasks involving multiple blades, coaxial mating, and gear meshing. From an engineering perspective, training in simulation on standard hole geometries can provide policies that are directly applicable to multiple real-world tasks, reducing the need to retrain policies or design separate search motions for each type of part.

Future work will build on the validated insertion skills to support a more complete autonomous workflow. Vision-based localization could automatically estimate and update the target pose as part positions change, complementing force-feedback corrections after contact and reducing reliance on prior hole-pose calibration. Integrating autonomous grasping, transfer, task-status assessment, and failure recovery with the insertion policy could support continuous operation and autonomous retries after failure. Building on these capabilities, we will explore insertion tasks with a wider range of mating structures and operational requirements, working toward an autonomous closed-loop system that covers the full insertion process.

\hypertarget{conclusions}{%
\section{Conclusions}\label{conclusions}}

The proposed framework learns contact-rich precision insertion entirely in simulation and deploys the resulting policies directly on real robots. The framework combines target poses with compact three-dimensional fingertip force feedback, uses a decoupled gated reward to coordinate alignment and insertion, and stabilizes learning from force feedback through force-signal smoothing and state-independent standard deviations. The resulting policies perform insertion across multiple hole geometries with a minimum nominal clearance of 0.02 mm, without real-world demonstrations or fine-tuning. Simulation and real-robot results show that force feedback helps the policies correct motion under localization errors, improving insertion success while reducing peak contact forces.

The policies also generalize across clearances and geometries and to unseen real-world insertion tasks. In the ManipulationNet evaluation, five policies trained on Tol. III for their respective hole geometries each cover the four real-world clearance levels of that geometry. The system achieved the first perfect score of 20/20 under the Human-in-the-Loop protocol, in which insertion motions are executed autonomously and the human operator only reports task status. A single Hexagon-III policy achieves an overall success rate of 95.0\% across the eight unseen real-world insertion tasks. These results demonstrate that learning entirely in simulation can produce precision insertion skills that can be deployed directly and reused across tasks, providing an effective solution for robotic assembly of diverse components.

\bibliographystyle{IEEEtran}
\bibliography{engineering_references}

\clearpage
\twocolumn[\section*{Supplementary Material}]
\begingroup
\setcounter{section}{0}
\setcounter{figure}{0}
\setcounter{table}{0}
\renewcommand{\thesection}{S\arabic{section}}
\renewcommand{\thesubsection}{S\arabic{section}.\arabic{subsection}}
\renewcommand{\thefigure}{S\arabic{figure}}
\renewcommand{\thetable}{S\arabic{table}}
\renewcommand{\theequation}{S\arabic{equation}}
\setcounter{equation}{0}
\normalsize
\hypertarget{additional-details-of-policy-learning-and-control}{%
\section{Additional details of policy learning and control}\label{additional-details-of-policy-learning-and-control}}

\hypertarget{observations-actions-and-compliant-tracking}{%
\subsection{Observations, actions, and compliant tracking}\label{observations-actions-and-compliant-tracking}}

The policy input in Section 2.1 of the main text consists of pose, motion, and fingertip force information available on the real robot. The force-feedback policy receives 26 observation components in the following order: relative target position (3), end-effector orientation quaternion (4), relative target orientation quaternion (4), end-effector linear and angular velocities (3 each), three-dimensional fingertip force, and the previous six-dimensional smoothed action. The pose-only policy omits the three force components and retains the other 23 inputs. The relative position describes the spatial relationship between the current end effector and the target. The end-effector orientation specifies the robot's current orientation, whereas the relative orientation describes its relationship to the target orientation. Training and deployment use the same quaternion component order and reference orientation, and deployment retains the observation normalization statistics learned during training. The critic additionally receives part poses, robot state, and randomized dynamics parameters for value estimation.

The policy output is clipped and then smoothed over time:

\[
\tilde{\mathbf a}_t=\beta\,\operatorname{clip}(\mathbf a_t,-1,1)
+(1-\beta)\tilde{\mathbf a}_{t-1},
\tag{S1}
\]
where \(\beta=0.2\) is the action smoothing coefficient, distinct from the force smoothing coefficient \(\alpha\) in Eq. (6) of the main text. The first three smoothed components are scaled to give translation increments relative to the current end-effector position. The remaining three components are scaled to form an axis-angle rotation increment, which is composed with the current orientation. The target roll and pitch are then fixed to keep the gripper pointing downward, while yaw adjustments are retained. Thus, both the network output and the action history are six-dimensional, but execution has four effective degrees of freedom: three translations and yaw. The translation and rotation scales in simulation are 0.02 m and 0.097 rad per axis, respectively. Smaller command scales are used on the real robot, as described in Section S4.

The dead zone in Eq. (2) of the main text is applied separately to each component of the six-dimensional Cartesian control vector. A component is set to zero when its magnitude does not exceed the threshold; otherwise, the threshold is subtracted from its magnitude while preserving its sign. Randomization therefore affects both contact friction and whether a small control correction produces motion. The robot's redundant posture is regulated through a null-space torque \cite{khatibOperationalSpace1987}:

\[
\begin{aligned}
\mathbf N_\tau&=\mathbf I-\mathbf J^{\mathsf T}
(\mathbf J\mathbf M^{-1}\mathbf J^{\mathsf T})^{-1}\mathbf J\mathbf M^{-1},\\
\boldsymbol\tau_{\mathrm{null}}&=\mathbf N_\tau\mathbf M
\left[\mathbf K_{p,n}(\mathbf q_0-\mathbf q)-\mathbf K_{d,n}\dot{\mathbf q}\right].
\end{aligned}
\tag{S2}
\]

In Eq. (S2), \(\mathbf M\) is the joint-space inertia matrix; \(\mathbf q\) and \(\dot{\mathbf q}\) are the joint positions and velocities; \(\mathbf q_0\) is the reference joint configuration; \(\mathbf K_{p,n}\) and \(\mathbf K_{d,n}\) are the null-space stiffness and damping matrices; \(\mathbf I\) is the identity matrix; and \(\mathbf N_\tau\) is the torque-space projection matrix. When the Jacobian has full row rank, \(\mathbf J\mathbf M^{-1}\mathbf N_\tau=\mathbf 0\), so this regulation does not affect acceleration in the primary task directions under the instantaneous dynamics.

\hypertarget{geometric-rewards-and-insertion-stages}{%
\subsection{Geometric rewards and insertion stages}\label{geometric-rewards-and-insertion-stages}}

The geometric errors in Eqs. (3)--(5) of the main text are computed from the simulated part poses. The planar distance \(d_{xy}\) and vertical distance \(d_z\) are measured between the peg reference point and its final target. The yaw error \(d_R\) is the smallest angular difference from the target orientations allowed by the task. Each angular difference is first mapped to \([-\pi,\pi)\) and then converted to its absolute value. As described in the main text, the circular task omits the yaw reward and gate. Table S1 gives a representative configuration for the hexagonal task to illustrate the scales of the reward components.

\begin{table}[t]
\centering
\caption{Representative parameters of the decoupled gated reward.}
\label{tab:s1}
\scriptsize
\setlength{\tabcolsep}{2pt}
\begin{tabular}{@{}p{0.27\linewidth}p{0.34\linewidth}p{0.31\linewidth}@{}}
\toprule
Reward component & Parameters & Values \\
\midrule
Planar alignment & $(k_{xy},b_{xy}),\ w_{xy}$ & $(50,2),\ 2.5$ \\
Yaw alignment & $(k_R,b_R),\ w_R$ & $(50,2),\ 2.5$ \\
Vertical insertion & $(k_z,b_z),\ w_z$ & $(50,1),\ 2.5$ \\
Alignment gating & $\kappa,\ \epsilon_R$ & $100\ \mathrm{m}^{-1},\ 1^\circ$ \\
Action magnitude and change & $\lambda_a,\lambda_{\Delta a}$ & $0.05,\ 0.1$ \\
Insertion-stage rewards & $w_e,w_h,w_s$ & All equal to 1 \\
\bottomrule
\end{tabular}
\end{table}

Note: Position and angular errors are measured in m and rad, respectively; \(k_{xy}\), \(k_z\), and \(k_R\) have the corresponding inverse units. The roles of these parameters are described in Section 2.2 of the main text.

To define the insertion stages, let \(z_{\mathrm{disp}}\) denote the signed vertical displacement of the peg reference point from its final target, positive when the peg is above the target. The stage indicators are

\[
\begin{aligned}
I_{j,t} = \mathbb I[z_{\mathrm{disp},t}<h_jH]
\mathbb I[d_{xy,t}<\epsilon_{xy}]\mathbb I[d_{R,t}<\epsilon_{\mathrm{succ}}],
\\ \qquad j\in\{\mathrm{eng},\mathrm{half},\mathrm{succ}\}.
\end{aligned}
\tag{S3}
\]

In Eq. (S3), \(H\) is the reference height of the fixed part in the task, \(h_j\) is the normalized threshold for the corresponding stage, \(\epsilon_{xy}=2.5\) mm is the planar alignment threshold, and \(\epsilon_{\mathrm{succ}}\) is the task's angular success threshold. The last factor is omitted when orientation alignment is not required. Representative stage thresholds are 0.90, 0.55, and 0.04, respectively, with \(\epsilon_{\mathrm{succ}}=0.1\) rad for the hexagonal task. They correspond to approaching the insertion region, advancing further, and approaching the final target. An indicator equals 1 at every control step that satisfies its conditions, so deeper stages can also receive rewards from preceding stages. The action penalties in Eq. (5) of the main text apply to the six-dimensional smoothed action in Eq. (S1) and its change between consecutive steps.

\hypertarget{randomization-and-policy-optimization}{%
\subsection{Randomization and policy optimization}\label{randomization-and-policy-optimization}}

Observation randomization distinguishes calibration biases that remain constant within an episode from measurement noise that varies over time. The former are sampled at the beginning of each episode and applied throughout the episode to the hole position or end-effector orientation; the latter are resampled when observations are updated. Both use zero-mean Gaussian distributions. Orientation errors are applied through small rotations rather than by directly changing quaternion components. Dynamics parameters are sampled uniformly from their respective ranges at the beginning of each episode and remain fixed within that episode. Table S2 summarizes the representative settings and the evaluation ranges for the circular task in Fig. 4(c) of the main text. All methods compared on the same target task use the same settings.

\begin{table*}[t]
\centering
\caption{Observation and dynamics randomization.}
\label{tab:s2}
\scriptsize
\setlength{\tabcolsep}{2pt}
\begin{tabular}{@{}p{0.27\linewidth}p{0.34\linewidth}p{0.31\linewidth}@{}}
\toprule
Randomized quantity & Representative settings and L-shape baseline evaluation & Circle baseline evaluation \\
\midrule
Static hole position bias SD (mm) & 1.0 per axis & Same \\
Dynamic hole and end-effector position noise SD (mm) & 0.5 per axis & 1.0 per axis \\
Static end-effector orientation bias SD (rad) & $(0.08,0.08,0.01)$ & $(0.08,0.08,0.04)$ \\
Dynamic end-effector orientation noise SD (rad) & $(0.04,0.04,0.005)$ & $(0.04,0.04,0.04)$ \\
Linear/angular velocity noise SD & 0.01 m/s / 0.01 rad/s per axis & 0.02 m/s / 0.05 rad/s per axis \\
Dynamic fingertip force noise SD (N) & 0.1 per axis & Same \\
Part contact friction coefficient & $[0.3,0.6]$ & $[0.2,0.6]$ \\
Translational/rotational stiffness & $[100,300]$ N/m / $[30,50]$ N·m/rad & $[100,400]$ N/m / $[30,60]$ N·m/rad \\
Translational/rotational dead zone & $[0,0.2]$ N / $[0,0.04]$ N·m & $[0,0.5]$ N / $[0,0.1]$ N·m \\
\bottomrule
\end{tabular}
\end{table*}

Note: Orientation tuples are ordered as roll, pitch, and yaw; dynamic orientation noise is applied as an axis-angle perturbation. Gaussian noise values are standard deviations (SD), and square brackets indicate uniform sampling ranges.

During randomized training with force observations, a single random scale is also applied to the three-dimensional force input for the entire episode, and the input is set to zero with a specified probability:

\[
\begin{aligned}
\mathbf f_t^{\mathrm{obs}} &= ms\left(\bar{\mathbf f}_t+\boldsymbol\eta_t\right),\\
&m\sim\operatorname{Bernoulli}(1-p),\quad s\sim\mathcal U(0,2).
\end{aligned}
\tag{S4}
\]
where \(\boldsymbol\eta_t\) is the force observation noise in Table S2, \(m\) is an episode-level retention mask, \(p=0.3\) is the dropout probability, and \(s\) is an episode-level scale factor. Scaling exposes the policy to force inputs of different magnitudes, while dropout also teaches it to use pose and motion information when force information is unavailable. Both operations change the policy input without changing physical contact in simulation. They are disabled during normal force-feedback evaluation and real-world deployment; the force-zeroing ablation in Section S3 changes the force input separately.

The policy uses a two-layer LSTM with 1024 hidden units per layer and fully connected layers of widths 512, 128, and 64. Recurrent states are reset at the start of each episode. PPO \cite{schulmanProximalPolicyOptimization2017} uses a discount factor of 0.995, a generalized advantage estimation coefficient of 0.95, and a clipping coefficient of 0.2. Each environment collects 128 consecutive steps before an update, and each batch undergoes four optimization passes. The initial learning rate is \(10^{-4}\) and is adjusted according to the KL divergence of policy updates. The actor learns deployable actions from noisy observations, while the critic uses more complete simulation states to estimate returns. The trained network and normalization statistics remain unchanged during evaluation and real-world execution.

\hypertarget{analysis-of-force-smoothing-and-exploration-scale}{%
\section{Analysis of force smoothing and exploration scale}\label{analysis-of-force-smoothing-and-exploration-scale}}

\hypertarget{effects-of-ema-on-short-term-fluctuations}{%
\subsection{Effects of EMA on short-term fluctuations}\label{effects-of-ema-on-short-term-fluctuations}}

Equation (6) of the main text applies EMA to the force signal. To analyze its effect on rapid random fluctuations, consider an interval over which the contact force is approximately constant. Write the baseline-subtracted input as \(\mathbf f_t=\mathbf f^*+\boldsymbol\nu_t\), where \(\mathbf f^*\) is the true force over that interval and \(\boldsymbol\nu_t\) is a temporally independent, zero-mean perturbation before filtering, with covariance \(\mathbf C_\nu\). Neglecting the initialization transient, the smoothing error \(\mathbf e^f_t=\bar{\mathbf f}_t-\mathbf f^*\) satisfies

\[
\operatorname{Cov}(\mathbf e^f_t)
=\alpha^2\sum_{k=0}^{\infty}(1-\alpha)^{2k}\mathbf C_\nu
=\frac{\alpha}{2-\alpha}\mathbf C_\nu.
\tag{S5}
\]

For \(\alpha=0.25\), the smoothing error covariance is \(1/7\) of the input perturbation covariance. Meanwhile, EMA assigns a weight of \(\alpha(1-\alpha)^k\) to each past sample, giving a mean lag of

\[
\sum_{k=0}^{\infty}k\alpha(1-\alpha)^k
=\frac{1-\alpha}{\alpha}=3.
\tag{S6}
\]

The index \(k\) counts sampling updates before the current time, so the value 3 is measured in filter update steps. EMA trades some response delay for suppression of rapid fluctuations. When contact force continues to change during exploratory motion, the smoothed signal changes accordingly. These relations explain how the smoothing coefficient affects noise attenuation and response speed; Fig. 4(b) of the main text tests its effect on learning during contact.

\hypertarget{state-independent-standard-deviations-and-observation-perturbations}{%
\subsection{State-independent standard deviations and observation perturbations}\label{state-independent-standard-deviations-and-observation-perturbations}}

The Gaussian policy in Eq. (7) of the main text uses learnable, state-independent standard deviations. Let \(\ell_j=\log\sigma_{\theta,j}\), giving the covariance matrix \(\boldsymbol\Sigma=\operatorname{diag}(\exp(2\ell_1),\ldots,\exp(2\ell_d))\). With the policy parameters fixed, the KL divergence between the distributions for two observations, \(\mathbf o\) and \(\mathbf o+\Delta\mathbf o\), is

\[
D_{\mathrm{KL}}\!\left[\pi_\theta(\cdot\mid\mathbf o)\,\|\,
\pi_\theta(\cdot\mid\mathbf o+\Delta\mathbf o)\right]
=\frac12\Delta\boldsymbol\mu^{\mathsf T}
\boldsymbol\Sigma^{-1}\Delta\boldsymbol\mu,
\tag{S7}
\]
where \(\Delta\boldsymbol\mu=\boldsymbol\mu_\theta(\mathbf o+\Delta\mathbf o)-\boldsymbol\mu_\theta(\mathbf o)\). If the perturbation is sufficiently small and the mean function is locally differentiable, then \(\Delta\boldsymbol\mu\approx\mathbf J_\mu\Delta\mathbf o\), where \(\mathbf J_\mu\) is the Jacobian of the action mean with respect to the observation. The observation perturbation thus changes the policy distribution through the action mean, while the covariance remains the same for both observations. Force feedback can therefore still change the direction of motion correction without additionally changing the sampling scale in the same forward pass.

With state-dependent standard deviations, a change in observation also changes \(\ell_j(\mathbf o)\) and directly changes the conditional differential entropy in Eq. (8) of the main text through \(\sum_j\ell_j(\mathbf o)\). The state-independent parameterization removes this direct dependence, while its exploration scale continues to be learned through policy optimization. Specifically, the log-likelihood gradient with respect to the \(j\)th log standard deviation is

\[
\frac{\partial\log\pi_\theta(\mathbf a\mid\mathbf o)}{\partial\ell_j}
=\frac{(a_j-\mu_{\theta,j}(\mathbf o))^2}{\sigma_{\theta,j}^{2}}-1.
\tag{S8}
\]

The variable \(a_j\) is the \(j\)th action component. This gradient contributes to updates through advantage weighting and clipping in PPO, allowing the exploration scale in each dimension to adapt to training experience rather than being predicted directly from the current force observation. These relations describe the six-dimensional policy distribution before clipping, smoothing, and orientation constraints.

EMA and state-independent standard deviations act on the input and exploration, respectively. The former reduces short-term force fluctuations; the latter retains an observation-dependent action mean while making the exploration scale a learnable parameter shared across states. The learning improvements from combining the two, shown in Fig. 4(b) and Table 2 of the main text, are consistent with these effects.

\hypertarget{simulation-evaluation-and-additional-results}{%
\section{Simulation evaluation and additional results}\label{simulation-evaluation-and-additional-results}}

\hypertarget{reward-ablation-and-learning-metrics}{%
\subsection{Reward ablation and learning metrics}\label{reward-ablation-and-learning-metrics}}

Figure 4(a) of the main text compares rewards with force observations, and Fig. S1 provides results for the same task without force observations. The keypoint reward uses four pairs of corresponding points along the insertion axis and measures pose error as the mean distance between corresponding points:

\[
\bar d_{\mathrm{kp}}=\frac{1}{K}\sum_{k=1}^{K}
\|\mathbf p_k-\mathbf p_k^*\|_2,\qquad K=4,
\tag{S9}
\]
where \(\mathbf p_k\) and \(\mathbf p_k^*\) are the \(k\)th keypoints on the peg and its target, respectively. The keypoints span 0.15 m along the insertion axis. Three shaping terms of the form in Eq. (3) of the main text are applied to \(\bar d_{\mathrm{kp}}\), with \((k,b)\) set to \((5,4)\), \((50,2)\), and \((100,0)\), respectively \cite{narangFactoryFastContact2022,noseworthyFORGEForceguidedExploration2025}. Noncircular tasks include a yaw reward and apply yaw gating to the finest keypoint term. Action penalties and insertion-stage rewards are kept consistent with the corresponding comparison settings.

Curriculum learning first raises the target vertically by 20 mm and trains with this shallower target for approximately 400 epochs (Curr. I). Training then continues from the resulting policy for approximately another 400 epochs with the final insertion target restored (Curr. II). The decoupled reward and direct keypoint reward both use the final target from the start and train for approximately 800 epochs. In Fig. S1, the final-window success rates are 0 for the direct keypoint reward, 45.9\% for the second curriculum stage, and 70.8\% for the decoupled reward. Even without force observations, the decoupled reward guides the learning of alignment and insertion, with a final-window success rate approximately 25 percentage points above that of the two-stage curriculum.

\begin{figure}[t]
\centering
\includegraphics[width=0.98\columnwidth]{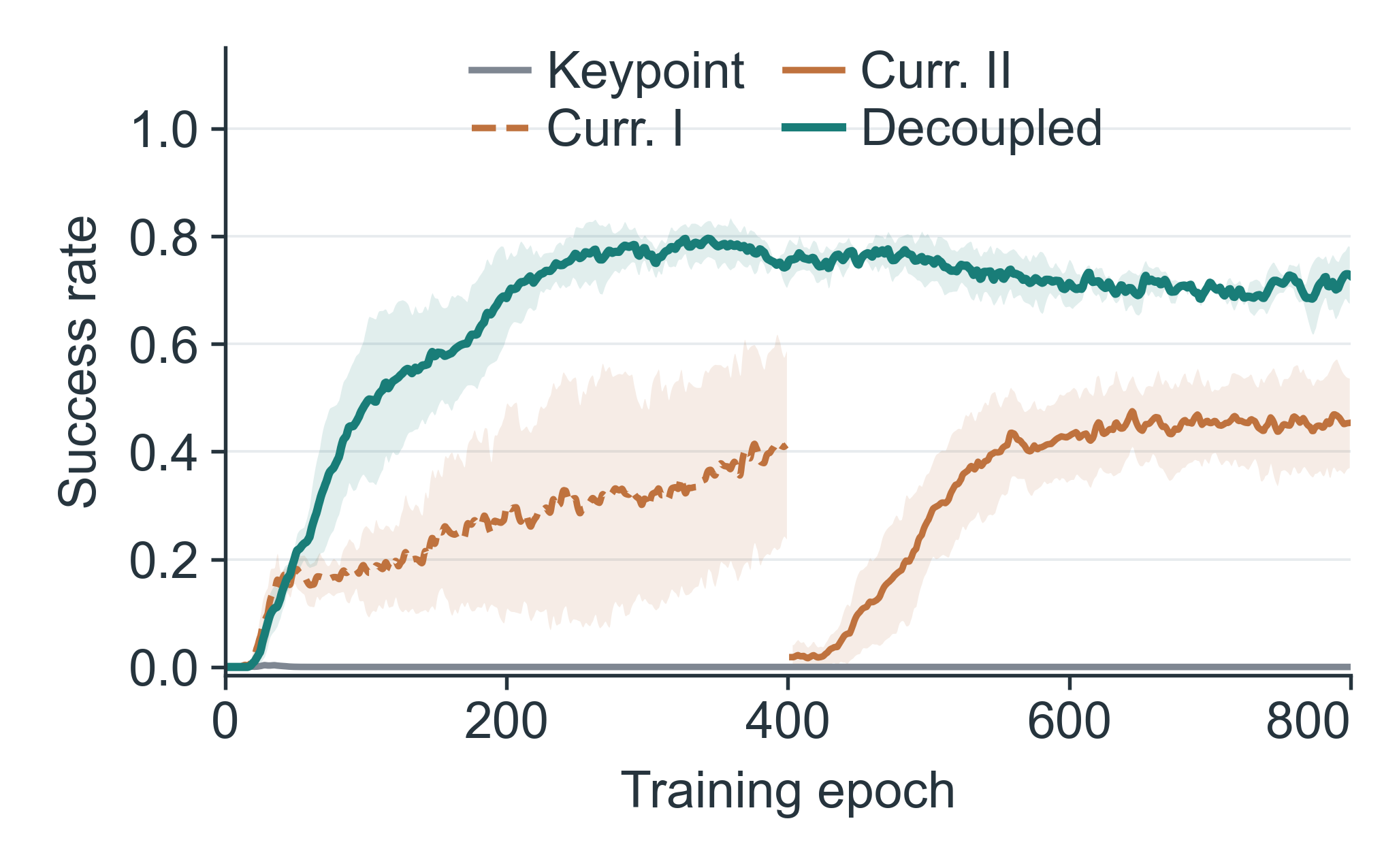}
\caption{Reward ablation on Hexagon-III without force observations. Keypoint denotes the keypoint setting with an additional yaw reward, and Decoupled denotes the decoupled gated reward. Curr. I and Curr. II are consecutive training stages with the target raised and restored, respectively; success in the first stage is measured against its shallower target. Curves and shaded regions show the mean and standard deviation across three independent training runs. Interpolation and display smoothing follow Fig. 4 of the main text; numerical comparisons use records before display smoothing.}
\label{fig:figs01}
\end{figure}

Figure S2 further explains how the two rewards guide learning differently. When the peg and target have the same orientation and differ only in translation, Eq. (S9) reduces to \(\bar d_{\mathrm{kp}}=\sqrt{d_{xy}^2+d_z^2}\). Planar and vertical errors are thus coupled in a single distance: reducing \(d_z\) can increase the keypoint shaping reward even before the peg is aligned with the hole opening. When \(d_z\) is large, a small planar correction has a relatively weak effect on the total distance. A direction that reduces this distance may therefore differ from an effective insertion direction constrained by contact with the hole opening.

During training with the keypoint reward, we observed behavior that departed from the task objective. When the policy failed to enter the tight-clearance hole, it instead placed the peg beside the hole body, bringing the peg reference point close to the final target height. Although planar error remained and insertion was not achieved, reducing the vertical error increased the keypoint shaping reward (Fig. S2(b)). This behavior illustrates reward hacking \cite{amodei2016concrete}: the policy improved the reward measure without completing the intended task. It shows how non-insertion states can provide favorable learning signals when insertion progress is measured solely by three-dimensional distance.

The decoupled reward retains an independent planar alignment reward throughout execution and modulates the vertical reward through \(\exp(-\kappa d_{xy})\) and yaw gating. For the same reduction in vertical error, a larger planar error yields a smaller vertical reward gain. Once alignment improves, this gain increases with the gate weight (Fig. S2(c)). The reward thus more explicitly encourages progress toward the target after improving alignment, while continuing to reward alignment corrections during advancement. The learning results in Fig. S1 and Fig. 4(a) of the main text test the effect of this design on policy learning.

\begin{figure*}[t]
\centering
\includegraphics[width=0.84\textwidth]{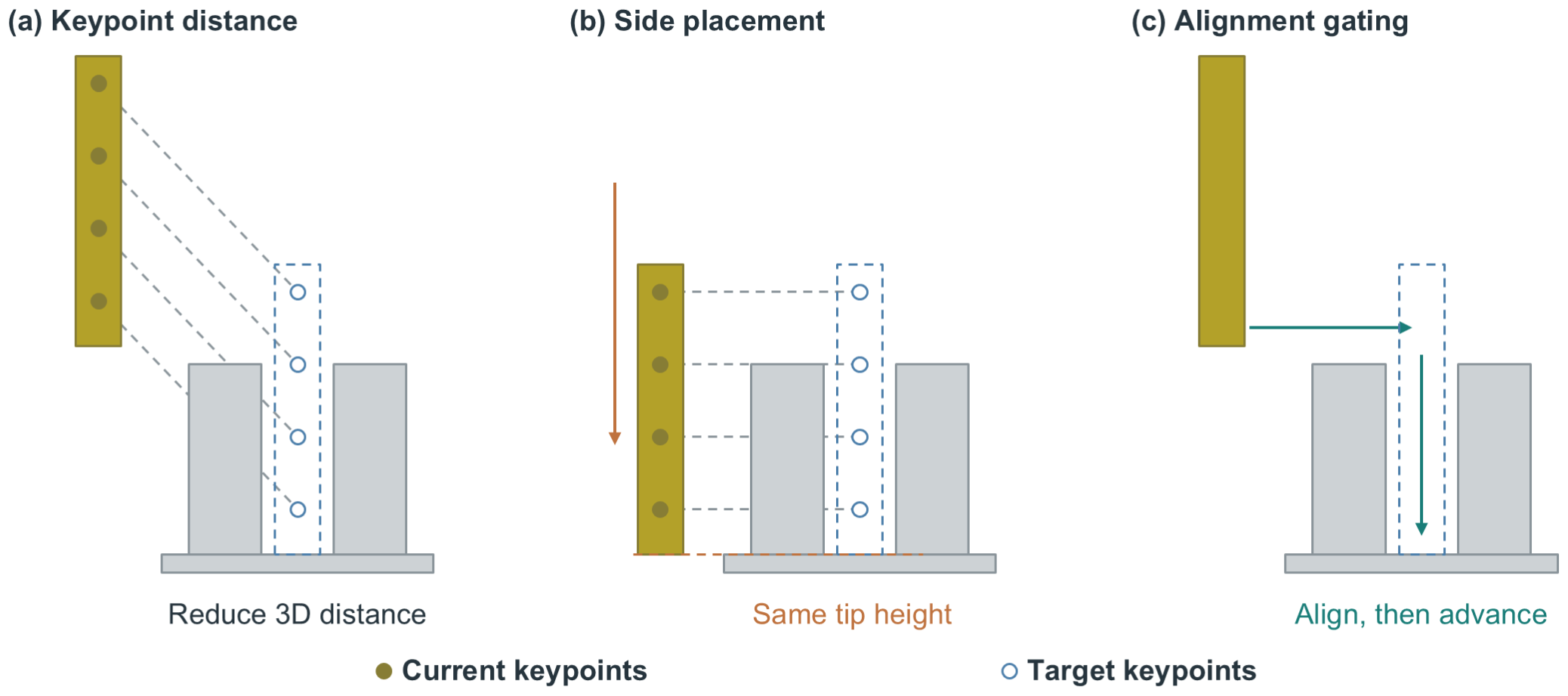}
\caption{The distinction between keypoint distance and effective insertion progress. (a) The keypoint reward combines planar and vertical errors between corresponding points into a three-dimensional distance. (b) Placement beside the hole body can reduce vertical error and keypoint distance without achieving insertion. (c) The decoupled reward retains a planar alignment term and increases the vertical reward as alignment improves. Filled and open circles in (a) and (b) denote current and target keypoints, respectively; the dashed blue outline denotes the target peg pose. This side-view schematic is not to scale in either geometry or keypoint spacing. Arrows indicate the motion directions discussed, not recorded policy trajectories.}
\label{fig:figs02}
\end{figure*}

Each reward-ablation setting uses three training seeds. The EMA and SI std. ablation on Square-IV likewise uses three training seeds, with 1000 training epochs. For plotting, the three curves within each setting are interpolated onto a common epoch grid over their shared epoch range. Each curve is then smoothed with an EMA span of 10 before the mean and population standard deviation are computed. This display smoothing is separate from the EMA applied to the force input and is not used to calculate the values in Table 2.

Suppose a seed has \(N\) equally spaced training records, with success rates \(s_1,\ldots,s_N\) at epochs \(e_1,\ldots,e_N\). The final-window success rate is the mean of the last 5\% of records. The normalized AUC and the first epoch at which success reaches 50\% are defined as

\[
\mathrm{AUC}_{\mathrm{norm}}=\frac1N\sum_{i=1}^{N}s_i,
\qquad
T_{50}=\min\{e_i:s_i\geq0.5\}.
\tag{S10}
\]

Each metric is computed separately for each training seed and then averaged across the three seeds. A run that never reaches 50\% has no \(T_{50}\). Only two seeds in the None setting of Table 2 reach this threshold, so a three-seed mean time to threshold is not reported. The final-window success rate reflects performance at the end of training, whereas the normalized AUC also captures how quickly and consistently success is achieved during training.

\hypertarget{classical-control-baselines-and-shared-evaluation-conditions}{%
\subsection{Classical control baselines and shared evaluation conditions}\label{classical-control-baselines-and-shared-evaluation-conditions}}

The control methods in Fig. 4(c) of the main text are compared across four tolerance levels for Circle and L-shape. Each method is evaluated on each target task with three evaluation seeds and 128 insertion trials per seed, for a total of 384 trials. Each trial lasts 20 s, and PPO executes the action mean. All methods on the same target task share the assets, initial pose randomization, observation noise, dynamics randomization, and success criteria. The first three classical methods use pose and velocity. Hybrid force/position control additionally uses the same baseline-subtracted, smoothed three-dimensional force as force-feedback PPO. True part states are used only to determine evaluation success and are not provided to these controllers.

Direct targets the observed hole pose and advances along the insertion direction after initial alignment, with the shared low-level compliant controller responding to contact. It serves as a reference for direct compliant insertion \cite{whitneyQuasiStaticAssembly1982}. Spiral and Raster add Archimedean spiral and reciprocating rectangular search trajectories, respectively, around the nominal hole center \cite{vanwykComparativePeginholeTesting2018,marvelMultiRobotAssembly2018}. The planar spiral offset is

\[
\mathbf p_{\mathrm{search}}(t)=
c\theta(t)\begin{bmatrix}\cos\theta(t)\\\sin\theta(t)\end{bmatrix},
\qquad \theta(t)=\omega t,
\tag{S11}
\]
where \(t\) is measured from the start of the search phase, \(c\) is the radial growth coefficient per unit angle, and \(\omega\) is the angular velocity. The search restarts when the radius reaches its limit. The rectangular search alternates direction between adjacent parallel scan lines and follows connecting segments to the next line. After covering the region, it retraces the path. All three pose-based methods first align to the nominal pose for 6 s, then begin advancement and the corresponding search.

Hybrid corrects the position reference according to the force error, and an inner compliant controller tracks that reference \cite{liHybridForcePosition2025}:

\[
\Delta\mathbf p_f=\mathbf A_f(\mathbf f_B-\mathbf f_d),
\qquad \mathbf p_{\mathrm{ref}}=\mathbf p_{\mathrm{nom}}+\Delta\mathbf p_f.
\tag{S12}
\]

In this force-to-position mapping, \(\mathbf f_B\) is the measured force transformed into the control frame, \(\mathbf f_d=[0,0,2]^{\mathsf T}\) N is the desired contact force, \(\mathbf A_f\) is a diagonal matrix of force-to-position coefficients, \(\mathbf p_{\mathrm{nom}}\) is the nominal advancing position, and \(\mathbf p_{\mathrm{ref}}\) is the corrected reference. The positive vertical direction is upward, so an axial contact force above the desired value shifts the reference upward. The implementation uses the three translational channels of the force-to-position mapping in \cite{liHybridForcePosition2025}, matching the available three-dimensional force observation; yaw is controlled through pose error. Initial pose alignment lasts 4 s, followed by advancement at the nominal speed with force-feedback corrections. Table S3 summarizes the fixed search and force/position parameters. None of the methods is tuned separately for individual tolerance levels.

\begin{table}[t]
\centering
\caption{Main settings of the classical control methods.}
\label{tab:s3}
\scriptsize
\setlength{\tabcolsep}{2pt}
\begin{tabular}{@{}p{0.22\linewidth}p{0.72\linewidth}@{}}
\toprule
Method & Search or feedback settings \\
\midrule
Direct & Advance toward the final target after aligning to the observed hole pose; no additional lateral search \\
Spiral & Maximum radius: 3 mm; four turns; angular velocity: 180°/s \\
Raster & Coverage: 6 × 6 mm; nine scan lines; path speed: 4 mm/s \\
Hybrid & Force-to-position coefficient: 1.25 mm/N per axis; lateral and vertical correction limits: 3 and 6 mm, respectively; nominal advancing speed: 8 mm/s \\
\bottomrule
\end{tabular}
\end{table}

Evaluation randomizes the end-effector position and yaw around the nominal initial pose, with position ranges of ±20, ±20, and ±10 mm along the three axes and a yaw range of ±0.2 rad. The initial hole yaw ranges are ±30° and ±22.5° for Circle and L-shape, respectively. Observation and dynamics randomization follow the corresponding settings in Table S2. All controllers produce the same form of six-dimensional normalized action, executed through shared action postprocessing and low-level compliant control. A trial is counted as successful once it reaches the required insertion depth and alignment conditions within the episode; satisfying the conditions repeatedly does not increase the count. Error bars in Fig. 4(c) show the population standard deviation across the three evaluation seeds.

\hypertarget{generalization-force-ablation-and-position-correction}{%
\subsection{Generalization, force ablation, and position correction}\label{generalization-force-ablation-and-position-correction}}

Figure 6 of the main text uses ten source policies trained on five hole geometries at Tol. III and Tol. IV. Each source policy is evaluated on all four tolerance levels of its own geometry and on all five geometries at its training tolerance. Removing the duplicated in-distribution condition for each source policy gives 80 source-policy--target-task combinations: 10 in-distribution, 30 cross-tolerance, and 40 cross-geometry combinations. Each combination is evaluated with three seeds, for a total of 384 trials. Figure 6(c) first computes the success rate for each combination and then averages combinations equally within each category, preventing a category's larger number of target tasks from changing its statistical weight.

Execution across tasks preserves the source policy's network parameters, normalization statistics, and observation definitions. When the target geometry changes, relative orientation inputs are still constructed using the reference orientation and quaternion representation from source-policy training. Success, however, is determined using the orientations allowed by the target geometry. This treatment represents the same alignment relationship consistently in the policy input while letting the target part determine whether insertion is complete. The force coordinate convention and EMA remain unchanged. Normal force-feedback evaluation disables the force dropout and random scaling used during training. All methods execute the action mean; evaluation seeds change environment initialization and random perturbations.

Both source and target tasks in Fig. 6(d) use Tol. III. Force and Force zeroed use the same force-feedback policy. Force zeroed sets the three-dimensional force input to zero before observation normalization, leaving all other observations and execution conditions unchanged. Pose-only uses an independently trained policy without force input. The first comparison tests how the same policy uses force feedback during execution; the latter compares transfer between policies trained with and without force inputs. Each condition likewise comprises 384 trials. The 95\% confidence intervals in Fig. 6(c) are obtained from 20,000 bootstrap resamples of target-task combinations. Figure 6(d) uses Wilson intervals based on 384 binary trial outcomes. These intervals describe uncertainty in the mean across task combinations and in the success proportion for each test condition, respectively.

Figure 5(b) of the main text separately examines motion correction under a fixed hole position bias. The L-shape-III task uses 36 nominal starting points on a 6 × 6 grid. The observed hole center is shifted by +2 mm in both \(x\) and \(y\), with no additional observation noise. Each starting point is executed for 150 control steps. The plotted distance reduction is the initial planar distance to the true hole center minus the final planar distance; positive values indicate movement toward the true center. The two policies use identical starting points and execution durations, so the displacement fields directly show their different responses to the same localization bias.

\hypertarget{real-world-deployment-and-experimental-procedures}{%
\section{Real-world deployment and experimental procedures}\label{real-world-deployment-and-experimental-procedures}}

\hypertarget{from-policy-output-to-physical-insertion}{%
\subsection{From policy output to physical insertion}\label{from-policy-output-to-physical-insertion}}

The experimental platform uses a Franka Emika robot arm and Paxini PX-6AXGEN3 fingertip force sensors. For single-hole and unseen insertion tasks, the part is held in the gripper before the trial. The robot is guided manually to record the end-effector reference pose at completed insertion, and a pre-insertion position is then set above it. During control, relative target observations are computed from this calibrated reference and the robot state. The policy uses three-dimensional fingertip force to correct motion near the nominal target.

Before each insertion, a force baseline is recorded while the part is not in contact with the target. Subsequent readings are baseline-subtracted, smoothed with EMA, and supplied to the policy using the coordinate convention from training. With the sensor mounting orientation used in these experiments, the raw sensor components \((f_x^{\mathrm{sens}},f_y^{\mathrm{sens}},f_z^{\mathrm{sens}})\) map to the policy force components \((f_y^{\mathrm{sens}},f_z^{\mathrm{sens}},-f_x^{\mathrm{sens}})\). The superscript \(\mathrm{sens}\) denotes the sensor frame. This fixed mapping is determined by the mounting orientation and remains unchanged across trials with the same mounting. Deployment uses the observation normalization statistics learned during training, without adding observation noise, force dropout, or random scaling.

The policy outputs the action mean at 15 Hz, while the Cartesian impedance controller in SERL \cite{luoSERLSoftwareSuite2024} tracks the target pose at 1 kHz. Real-world execution retains the action smoothing in Eq. (S1), with translation and rotation command scales of 0.01 m and 0.045 rad per axis, respectively. Per-step displacement, yaw, and the workspace around the target are constrained. Roll and pitch maintain the prescribed grasp orientation, while the policy continuously adjusts three-dimensional position and yaw. Recurrent states, action history, and force filter states are reset for every trial so that each insertion starts from an independent initial state.

\hypertarget{questions-addressed-by-the-real-world-experiments}{%
\subsection{Questions addressed by the real-world experiments}\label{questions-addressed-by-the-real-world-experiments}}

\begin{figure}[htbp]
\centering
\includegraphics[width=0.94\columnwidth]{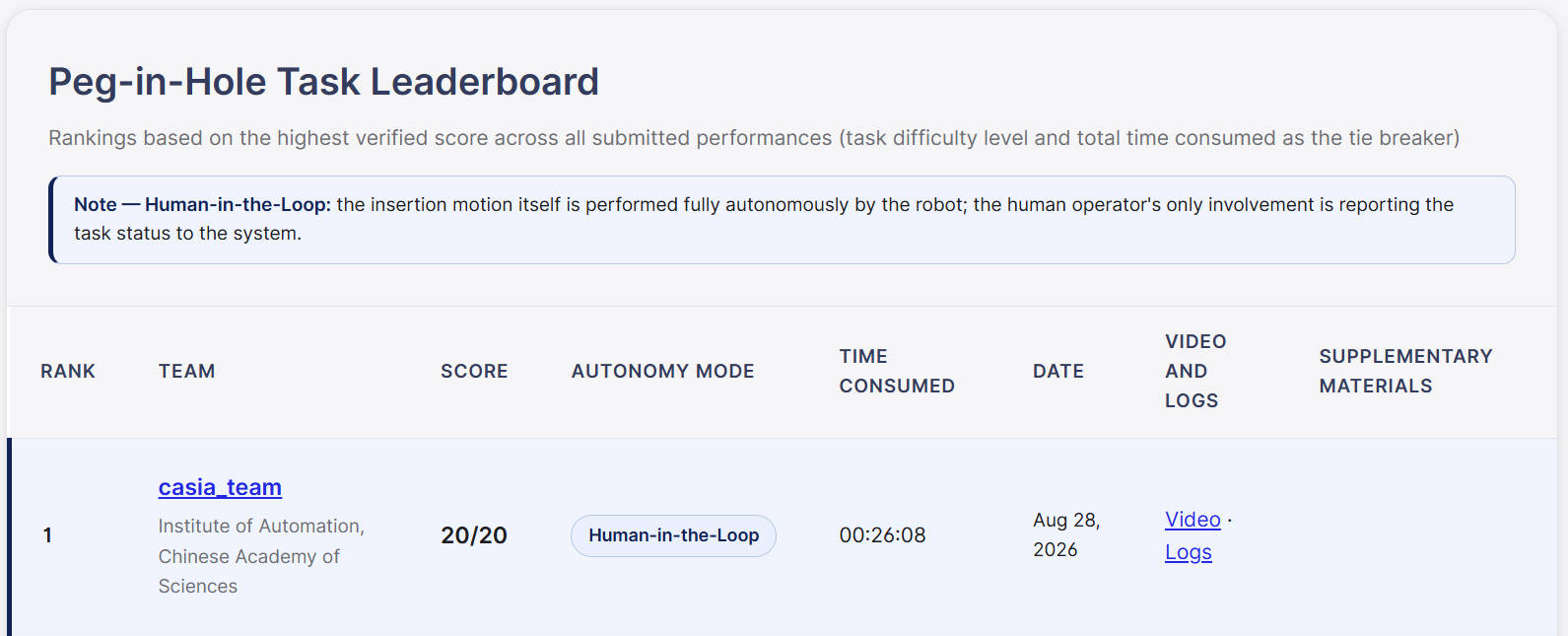}
\\[2pt]
\includegraphics[width=0.94\columnwidth]{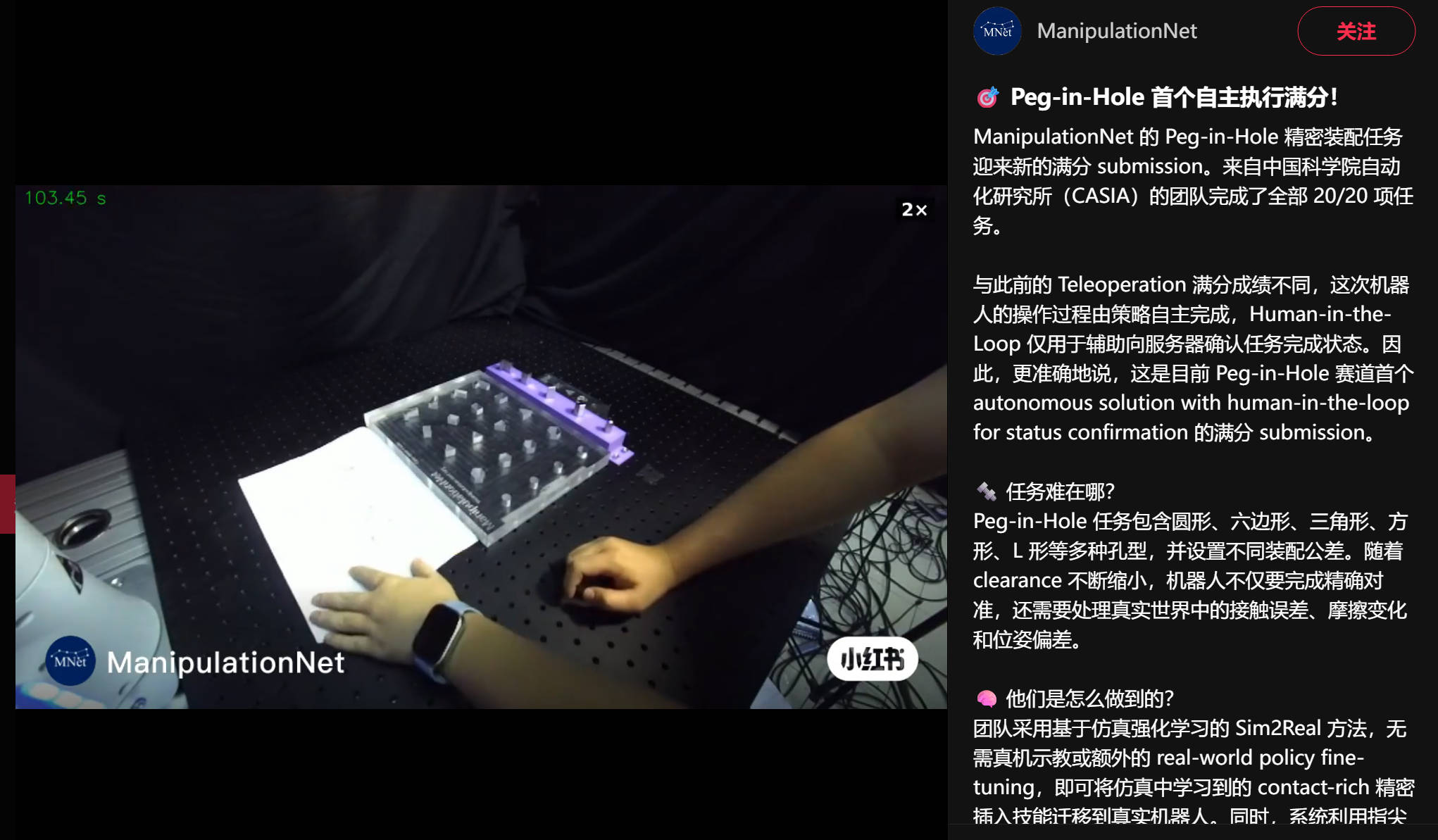}
\caption{Official record and public announcement of the ManipulationNet result. Top: the official leaderboard entry for our team (casia\_team), with a score of 20/20 under the Human-in-the-Loop protocol. Bottom: ManipulationNet's official social-media announcement recognizing our team as the first autonomous insertion solution to achieve a perfect score under this protocol. Sources: \href{https://manipulation-net.org/leaderboards/peg_in_hole.html}{official leaderboard} and \href{https://www.xiaohongshu.com/discovery/item/6aa2c3760000000028030276?source=webshare&xhsshare=pc_web&xsec_token=CBZMBEvFjMPJU7v_KvVKE_4bWnzdes_tuHofGBcmETLAs=&xsec_source=pc_share}{official social-media announcement}.}
\label{fig:s3}
\end{figure}

The single-hole experiments test insertion under tight fits. Circle, Hexagon, Triangle, and L-shape all use our custom Tol. IV parts, with 20 trials per policy per task. The initial position and yaw of the pregrasped peg are varied, without adding hole position bias. The hole position bias experiment further tests three levels of static planar bias on Hexagon-IV, with 15 trials per level. Force-feedback and pose-only policies use the same set of biased positions. The mean peak contact force in Table 3 is computed by first taking the peak force norm in each insertion and then averaging over the corresponding set of trials.

The ManipulationNet evaluation uses the official parts for five geometries and four tolerance levels. Each geometry uses only one policy trained at Tol. III in simulation to cover all four real-world tolerances of that geometry. Motion planning handles grasping and transfer, while the learned policy performs insertion. Human involvement under the Human-in-the-Loop protocol is limited to reporting task status and does not include control of the insertion motion. The 20/20 result in the main text means that all 20 official task items, comprising five geometries and four tolerances, were completed. Figure S3 shows the official leaderboard entry and ManipulationNet's public announcement of this result.

All eight unseen real-world insertion tasks use the same Hexagon-III force-feedback policy, with 20 trials per task. USB, two-pin and three-pin plugs, a DC power connector, gear meshing, and three electric-vehicle battery-circuit connectors together account for 160 insertions. When the object is changed, the corresponding insertion target is calibrated while the policy and its observation processing are retained, without policy retraining or fine-tuning. Figure 8 and Section 3.4 of the main text report results for each task. Representative insertions are shown in Supplementary Video 1, and consecutive single-hole L-shape trials with and without force feedback are shown in Supplementary Video 2.

\endgroup

\end{document}